\documentclass{article}

\usepackage[numbers]{natbib}
\usepackage[final]{nips_2016}
\usepackage[utf8]{inputenc} 
\usepackage[T1]{fontenc}    
\usepackage{url}            
\usepackage{booktabs}       
\usepackage{amsfonts}       
\usepackage{nicefrac}       
\usepackage{microtype}      
\usepackage[pdftex]{graphicx}
\usepackage{subcaption}
\usepackage{hyperref}       
\usepackage{float}

\title{Language Expansion In Text-Based Games}

\author{
  Ghulam Ahmed Ansari$^{+}$, Sagar J P$^{+}$, Sarath Chandar$^{*}$, Balaraman Ravindran$^{+}$\\
  ~\\ $^{+}$ Indian Institute Of Technology Madras, India.\\
   $^{*}$ University of Montreal, Canada.\\
}

\begin{document}

\maketitle

\begin{abstract}
Text-based games are suitable test-beds for designing agents that can learn by interaction with the environment in the form of natural language text. Very recently, deep reinforcement learning based agents have been successfully applied for playing text-based games. In this paper, we explore the possibility of designing a single agent to play several text-based games and of expanding the agent's vocabulary using the vocabulary of agents trained for multiple games. To this extent, we explore the application of recently proposed policy distillation method for video games to the text-based game setting. We also use text-based games as a test-bed to analyze and hence understand policy distillation approach in detail.
\end{abstract}

\section{Introduction}

Representation of data plays a major role in any machine learning task. Success of deep learning methods can be mainly attributed to their ability to learn meaningful task specific representations. While the initial success of deep learning was due to unsupervised pre-training \cite{bengio2007greedy}, \cite{hinton2006fast}, with the availability of large amount of labeled data and modern architectures and algorithms, end-to-end supervised training is the dominating strategy. However, Natural Language Processing (NLP) is one field in which pre-trained word embeddings\cite{mikolov2013distributed}, \cite{pennington2014glove} have been successfully used in several downstream techniques. While language modeling is the common task for learning word representations, human beings learn the language mainly by interaction.

Interaction based language learning has been recently explored in the context of learning to play text-based games \cite{lstmdqn} and dialogue based language learning \cite{weston2016dialog}. \cite{lstmdqn} proposed a Q-learning agent called LSTM-DQN which learnt to play simple Multi-User Dungeon (MUD) games by applying reinforcement learning. \cite{lstmdqn} also showed that by initializing the word embeddings of LSTM-DQN with embeddings pre-trained from a different game, one can achieve faster learning. However training a new agent for every new game has severe limitations. Even though one can transfer knowledge from previous games to the new game, it is still a one-way knowledge transfer.

The goal of this paper is to design a single agent which can learn to play all the games and have all-way knowledge transfer between these games. We achieve this by applying policy distillation method explored in \cite{DBLP:distillation} for DQN to LSTM-DQN model. Even though this is a straight-forward application of policy distillation, it is not guaranteed that policy distillation will work for LSTM-DQN architecture as different games have different vocabulary. We empirically verify that policy distillation indeed works for LSTM-DQN. Also, unlike video game agents, text-based game agents learn word embeddings in the representation layer which can be helpful for other tasks. In this paper, we show that policy distillation is the better way of expanding the language of one agent from several agents. We also provide deeper insights into how policy distillation functions using heat maps of the representation and control modules of the agent.

\section{Related Work}

\cite{lstmdqn} proposed a simple LSTM-DQN architecture for learning text based games. The action space in this architecture was restricted to be of one action word and one object word (e.g. \textit{go east}). \cite{he2015deep} considered games where the action space is also a natural language text. \cite{sukhbaatar2015mazebase} introduced mazebase, an environment for designing text based games and applied policy gradient methods with curriculum based training.

Deep Q-Networks \cite{dqnnature} have been initially proposed for learning arcade games. Ever since the proposal of DQN, it has been extended in several ways \cite{van2015deep,schaul2015prioritized,wang2015dueling}. \cite{DBLP:distillation} introduced a policy distillation approach for designing single agent that can play multiple games. This paper is an application of policy distillation method proposed in \cite{DBLP:distillation} for text-based games. Text-based games, unlike arcade video games, are interesting test-bed for understanding the policy distillation algorithm. Different games can have different vocabulary (with possible overlap in the vocabulary). Also it is easy to create simple text-based game environments with contradicting state dynamics which will shed some light on the effect of policy distillation.

\section{Multi-task Distillation Agent for learning representations from multiple sources}
\label{approach}

LSTM-DQN agent proposed in \cite{lstmdqn} differs from the standard DQN agent \cite{dqnnature} in two ways. Firstly, since the state is a sequence of words in the case of text-based games, LSTM-DQN uses an LSTM layer for state representation instead of a convolutional layer. This LSTM layer will have different vocabulary for different games. Following \cite{lstmdqn} a whole sentence is passed through the LSTM word by word and an output is generated by it for every word. The mean of all these outputs of LSTM is passed to the next fully connected layer of the network. Secondly, MUD-games use multi-word textual commands as actions. For each state in the game, we predict Q-values of all possible actions and all available objects using the same network. An average of the Q-values of the action $a$ and the object $o$ is used as a measure of the Q-value of the entire command $(a,o)$.

\begin{figure}[h]
\centering
\begin{subfigure}{.5\textwidth}
  \centering
  \includegraphics[width=0.9\linewidth]{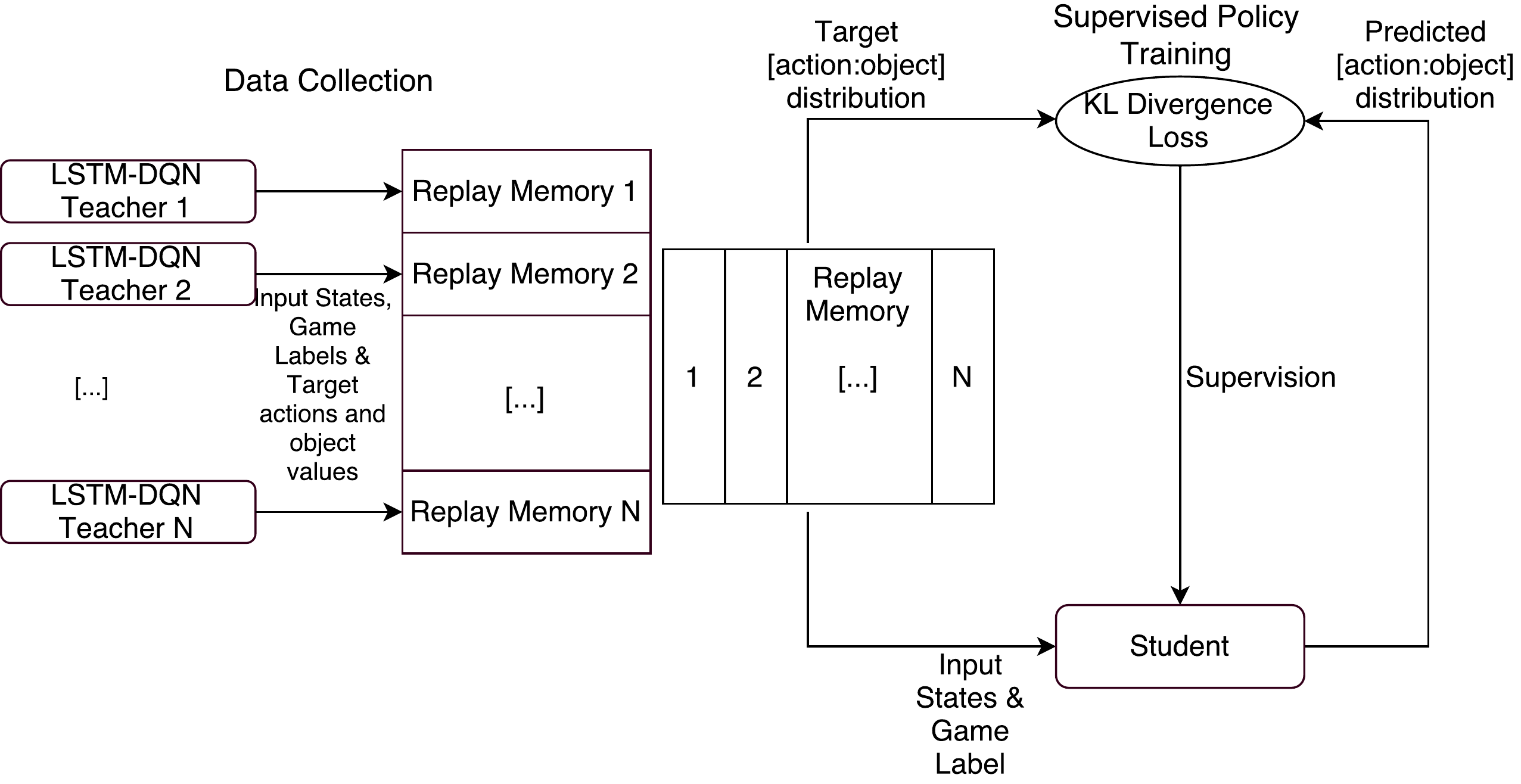}
  \caption{Multi Task Policy Distillation for Text-based games}
  \label{fig:Multitext}
\end{subfigure}\hfill
\begin{subfigure}{.3\textwidth}
  \centering
  \includegraphics[width=0.9\linewidth]{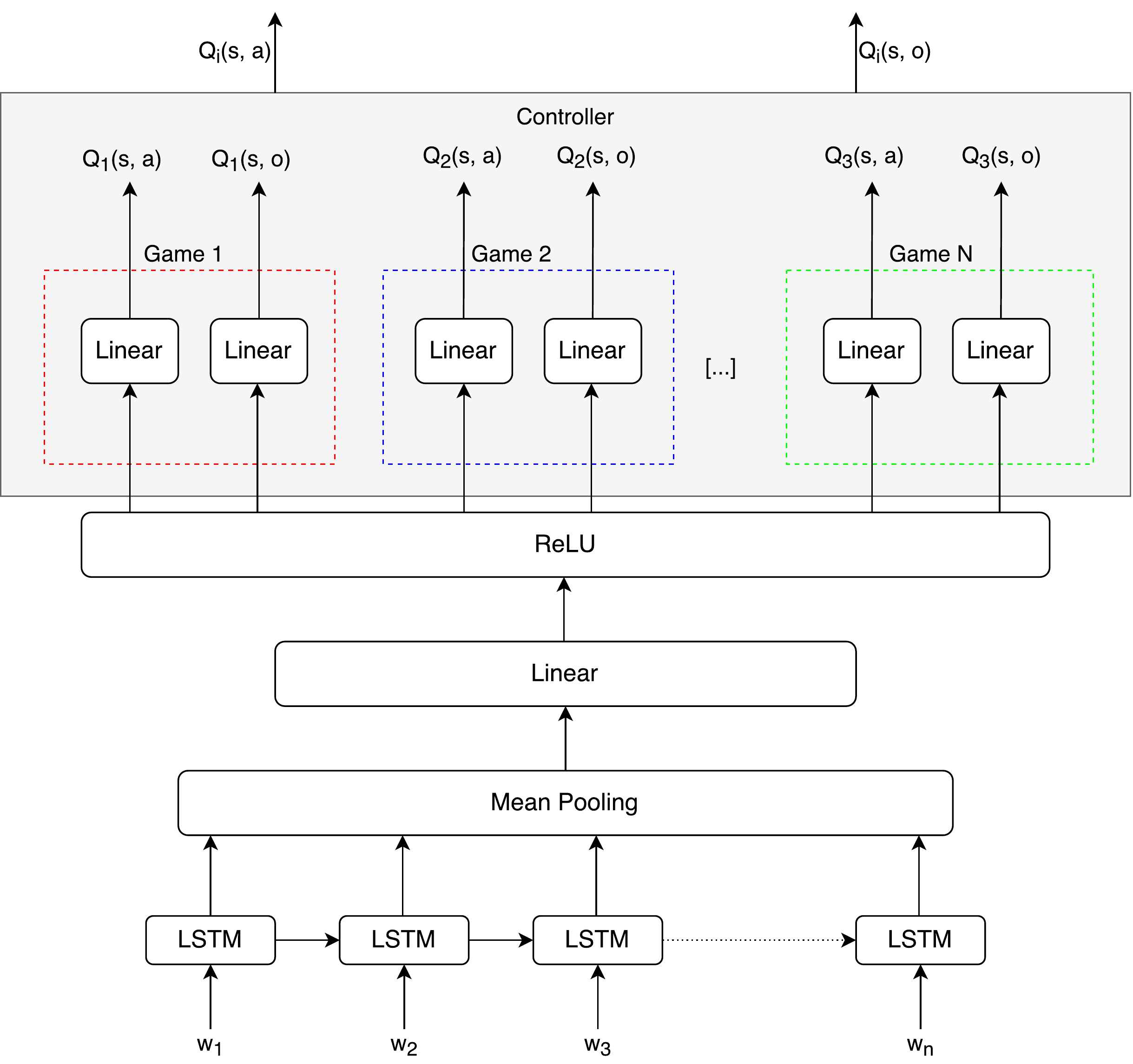}
  \caption{Student Network Architecture}
  \label{fig:Student}
\end{subfigure}
\caption{Learning representations from multiple text-based games}
\label{fig:ourmodel}
\end{figure}

Now we describe our method to learn language representations from multiple games with stochastic textual descriptions. First we train $n$ single-game LSTM-DQN teachers ($T$) separately. We want a single agent to be able to play all $n$ games, therefore we train a separate student neural network ($S$) to learn the optimal policies of each of these games. Each of the $n$ expert teachers produce a set of state-action and state-object values which we store along with inputs and game labels in a memory buffer. A separate memory buffer is maintained for each game as shown in Figure~\ref{fig:Multitext} and training happens sequentially, i.e training is done one by one, on samples drawn from a single game buffer at a time.

As we have $n$ different games and each game has to output an action and object at each time-step, the student network $S$ has a controller which can shift between $n$ different action, object output modules as shown in Figure~\ref{fig:Student}. The student network uses the input game label to switch between the corresponding modules.

The outputs of the final layer of the teachers are used as targets for the student $S$, after passing through a softmax function with a temperature parameter($\tau$). The targets to the student $S$ are thus given by $\mathrm{softmax}(\frac{\mathbf{q}^T}{\tau})$, where $\mathbf{q}^T$ is
the vector of Q-values of Teacher $T$. The outputs of the final layer of the student $S$ are also passed through a softmax function and can be given by $\mathrm{softmax}(\mathbf{q}^S)$, where $\mathbf{q}^S$ is
the vector of Q-values of Student $S$. KL divergence loss function is used to train the student network $S$.
\begin{equation}
    \mathcal{L}(X,\theta_{S}) = \sum_{i=1}^{|X|}softmax(\frac{\mathbf{q}^{T}_{i}}{\tau})ln \frac{softmax(\frac{\mathbf{q}^{T}_{i}}{\tau})}{softmax(\mathbf{q}^{S}_{i})}
\end{equation}
Where dataset $X= \{(s_i,\mathbf{q}_i)\}_{i=0}^{N}$ is generated by Teacher $T$, where
each sample consists of a short observation sequence $s_i$ and a vector
$\mathbf{q}_i$ of un-normalized Q-values with one value per action and one value per object. In simpler terms, $\mathbf{q}_i$ is a concatenation of the output action and object value function distributions. We call this agent as multi-task distillation agent. This is a straight-forward extension of distillation agent for video games proposed in \cite{DBLP:distillation} to text-based games.

For comparison, we also train a multi-task LSTM-DQN agent. For multi-task LSTM-DQN, the approach is similar to single-game learning: the network is optimized to predict the average discounted return of each possible action  given a small number of consecutive observations. Like in multi-task distillation, the game is switched every episode, separate replay memory buffers are maintained for each task, and training is evenly interleaved between all tasks. The game label is used to switch between different output modules as in multi-task policy distillation, thus enabling a different output layer, or controller, for each game. With this architecture in place, the multi-task LSTM-DQN loss function remains identical to single-task learning.

\section{Experiments and Analysis}
\label{setup}

In this section, we do an extensive analysis of policy distillation applied to LSTM-DQN. Specifically, we are interested in answering the following questions:

\begin{enumerate}
    \item How to learn multiple games with contradicting state dynamics?
    \item What is the mechanics of multi-task policy distillation?
    \item What can we say from the visualization of word embeddings?
    \item How to do transfer learning using policy distillation?
    \item How is the performance of policy distillation when compared to multi-task LSTM-DQN?
\end{enumerate}

We first explain the text-based game environment used in the experiments before reporting our findings.

\subsection{Game Environment}

We conduct experiments on 5 worlds which we have built upon the \textit{Home World} created by \cite{lstmdqn}\footnote{using Evennia (\url{http://www.evennia.com/})}. For our vocabulary expansion experiments, we have created multiple worlds, such that no two of the worlds have identical vocabulary. Out of the 5 worlds we created, game-$4$ and game-$5$ have a different layout. The games have a vocabulary of 90 words on an average. Every room has a variable set of textual descriptions among which one is randomly provided to the agent on entering that room. The quests (tasks given to the agent) and room descriptions are structured in a similar fashion to \cite{lstmdqn} in order to constrain an agent to understand the underlying information from the state so as to achieve higher rewards.

\begin{figure}[h]
    \centering
   \includegraphics[width=\linewidth]{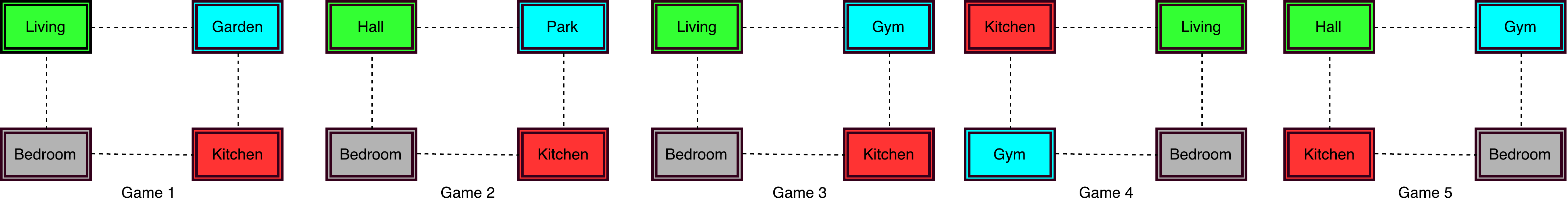}
   \caption{Different game layouts}
   \label{fig:Layout}
\end{figure}

Each of the 5 worlds consists of four rooms -- a bedroom, a living room, a garden, and a kitchen -- that can be connected in different ways as shown in Figure \ref{fig:Layout}. Each room has an object that the player can interact with. The player has to figure out the right room and the object to interact with based upon the given quest. "eat pizza", "go west" etc. are few  examples of possible interactions of the player with the environment. There is no stochasticity in transitions between the rooms. The start states are random and the player can spawn into any random room with a random quest provided to it. The state representation that the player has, contains a textual description of its current state and quest. The action set and quests that are given to the agent in all 5 games are the same. We have also added alternate descriptions for all the rooms in different games.

We use the same evaluation metrics and procedure as used in \cite{lstmdqn}. To measure an agent's performance we use the cumulative reward acquired per episode averaged over multiple episodes and the fraction of episodes in which the agent was able to complete the quest within $20$ steps during evaluation. The best value for average reward a player can get is $0.98$, since each step incurs a penalty of $-0.01$. In the worst case scenario the goal room can be diagonally opposite to the start room and so in order to obtain the best score the agent should be completing the given task within 2 to 3 steps without selecting any wrong command. Giving a negative reward for every step also enables the agent to optimize the number of steps.

\subsection{How to learn multiple games with contradicting state dynamics?}
In this section we analyze the performance of multi-task policy distillation agent (student) when the game state dynamics are contradicting. We use expert trajectories from games $1,2,4$ as teacher's to our student network. Game $1,2$ have same layout and game $4$ has a different layout as can be seen in Figure \ref{fig:Layout}. We first train a student network with the size of the first linear layer as 50.

\begin{figure}[H]
\centering
\begin{subfigure}{.5\textwidth}
  \centering
  \includegraphics[width=0.9\linewidth]{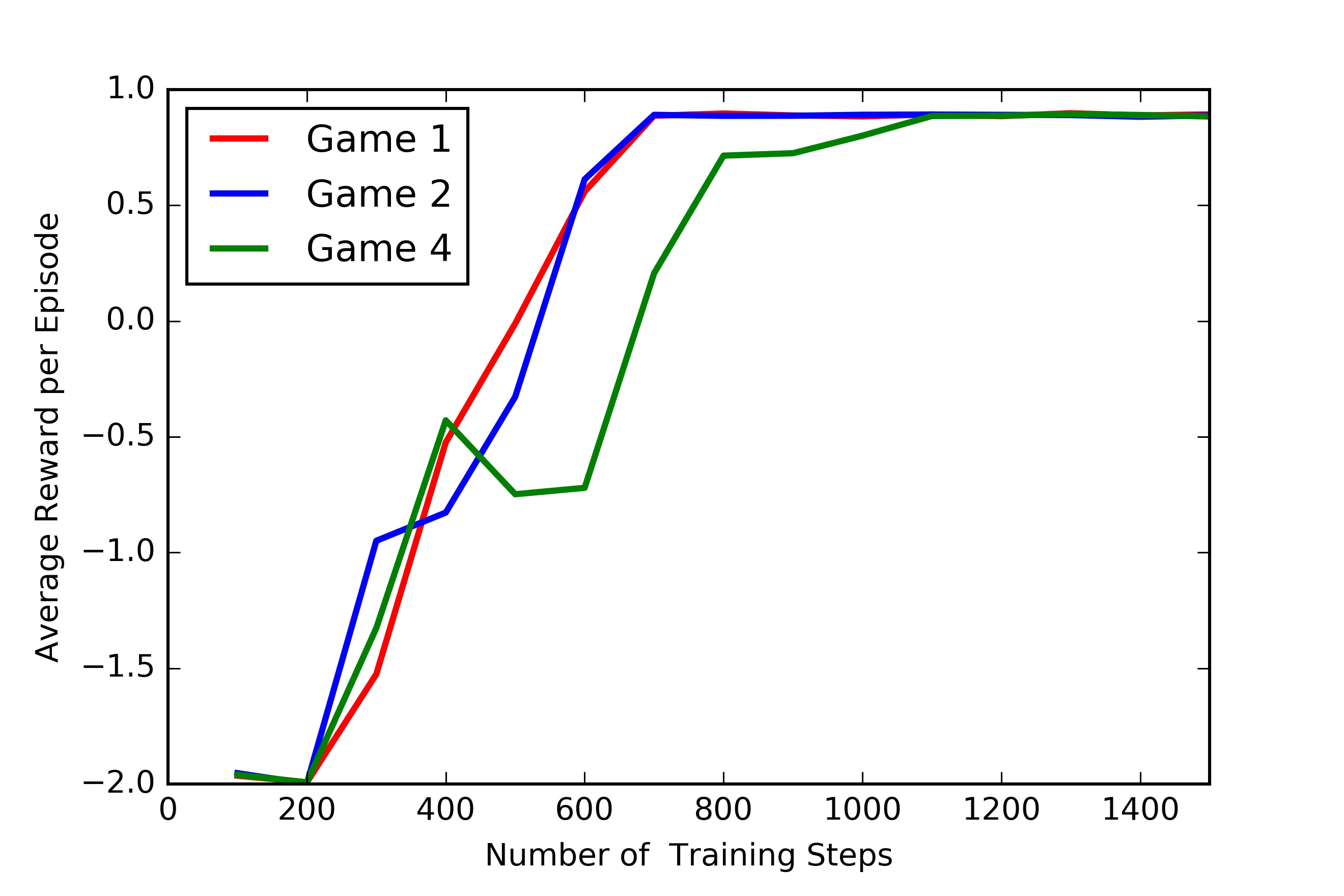}
  \caption{Average Reward per Episode}
  \label{fig:avgrs}
\end{subfigure}%
\begin{subfigure}{.5\textwidth}
  \centering
  \includegraphics[width=0.9\linewidth]{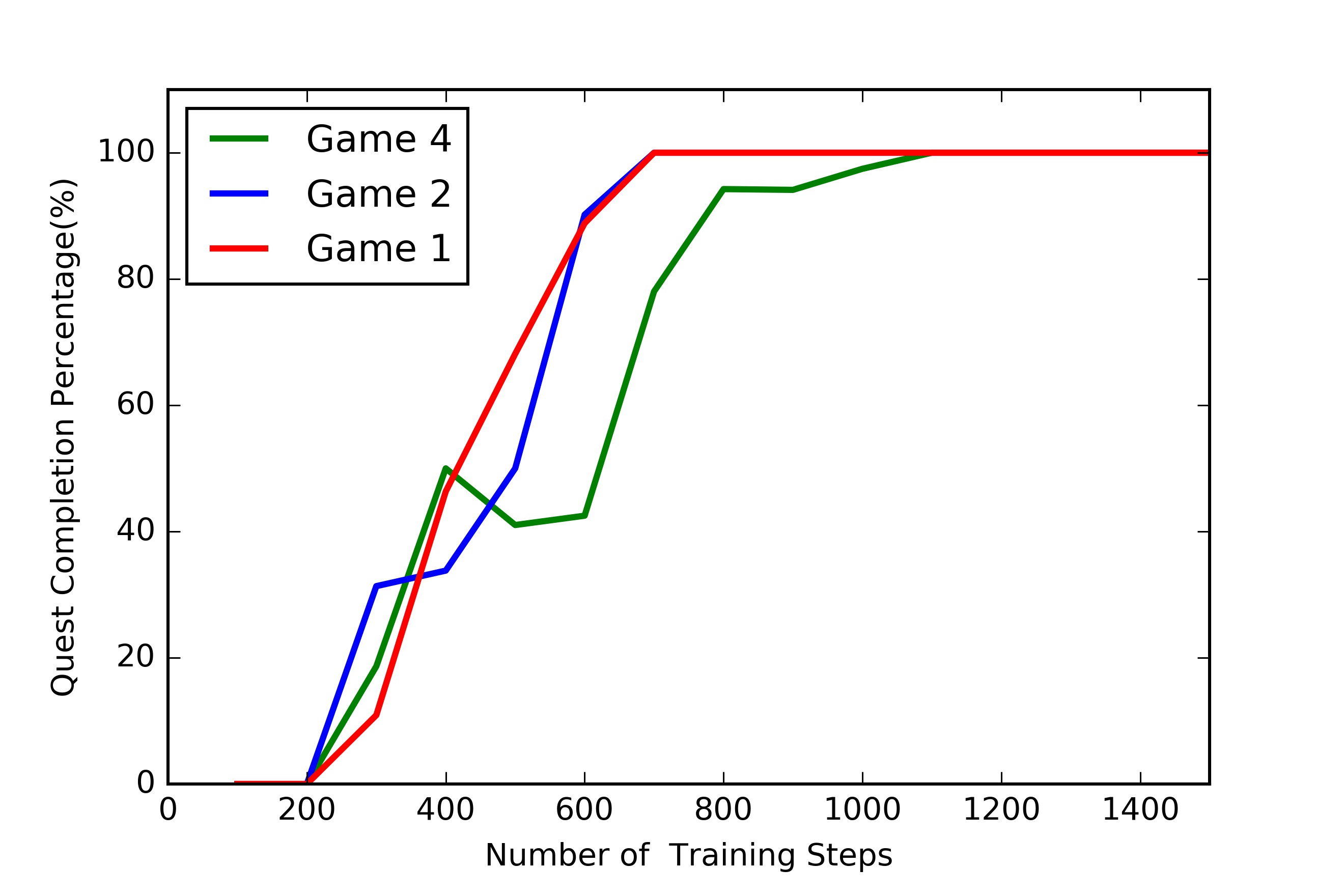}
  \caption{Quest Completion Percentage}
  \label{fig:qcomps}
\end{subfigure}
\caption{Training curves of smaller multi-task policy distillation agent (student) with $1^{st}$ linear layer size as $50$}
\label{fig:perfs}
\end{figure}

In Figure \ref{fig:perfs} we see that after around $1000$ training steps, the student is not only able to complete $100\%$ of the tasks given to it but additionally it also is able to get the best average reward possible on all the $3$ games. Note that the size of the $1^{st}$ linear layer for the teacher networks is $100$ which is double of that of the smaller student network used here. Even though student network had lesser number of parameters, it was able to learn all the games completely and was able to perform on par with the teacher networks eventually. This highlights the generalization capabilities of multi-task policy distillation method for text-based games.\label{smallernet}
An interesting observation that can be made from \ref{fig:avgrs} is that, initially for all the games the average reward increased but around 400 training steps the average reward for the game 4 started decreasing and after a while it started increasing again. Similar phenomenon can observed in the Figure \ref{fig:qcomps} too. One reason for this we suspect is that, the layout of game 4 is completely different compared to the other games. Due to this factor, an update in the network weights in the direction of game $1,2$ could negatively affect it's performance on game $4$. To strengthen our belief, we performed the same experiments with doubled size of the $1^{st}$ Linear layer.\label{larger}
\begin{figure}[htbp]
\centering
\begin{subfigure}{.5\textwidth}
  \centering
  \includegraphics[width=0.9\linewidth]{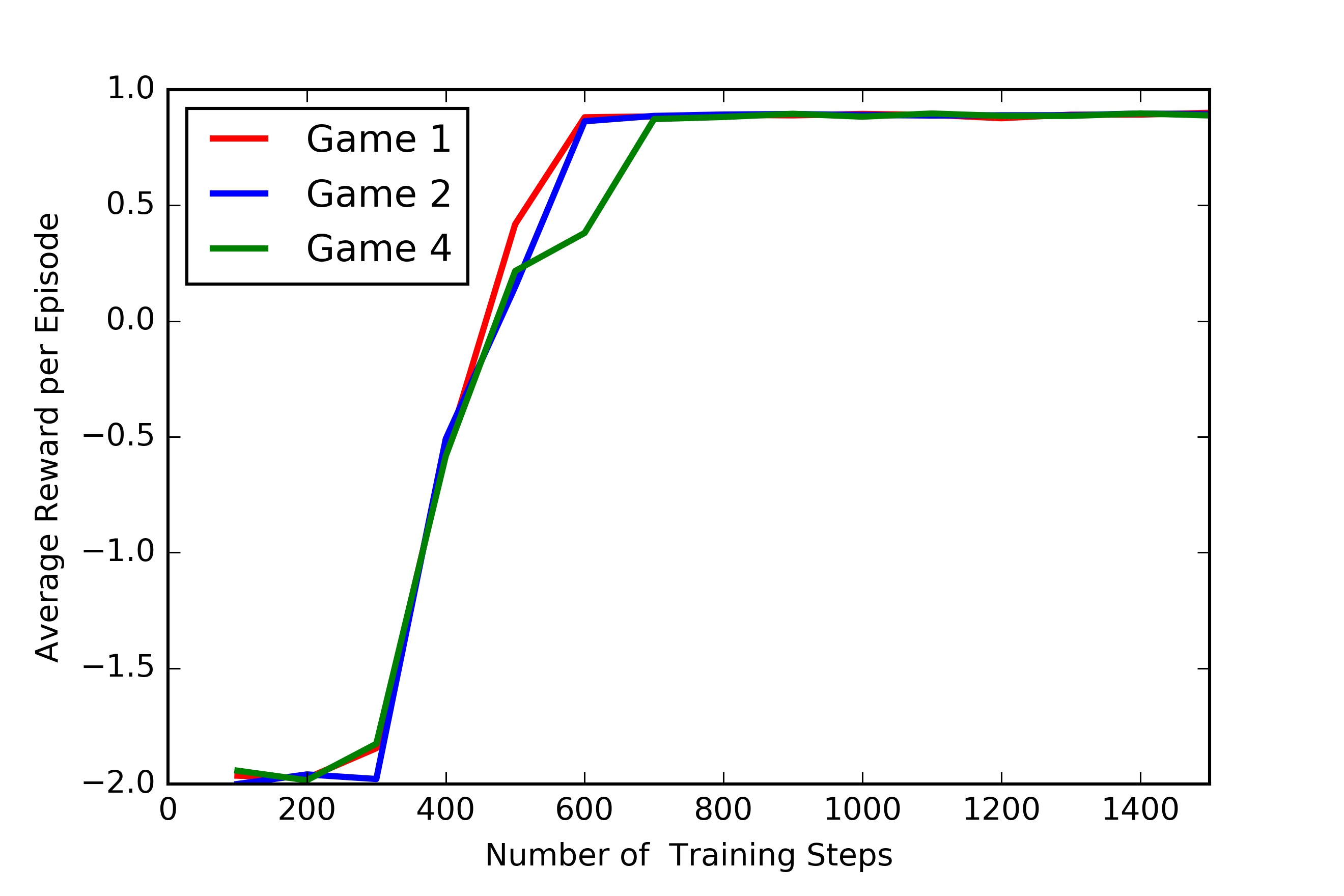}
  \caption{Average Reward per Episode}
  \label{fig:lsavgrs}
\end{subfigure}%
\begin{subfigure}{.5\textwidth}
  \centering
  \includegraphics[width=0.9\linewidth]{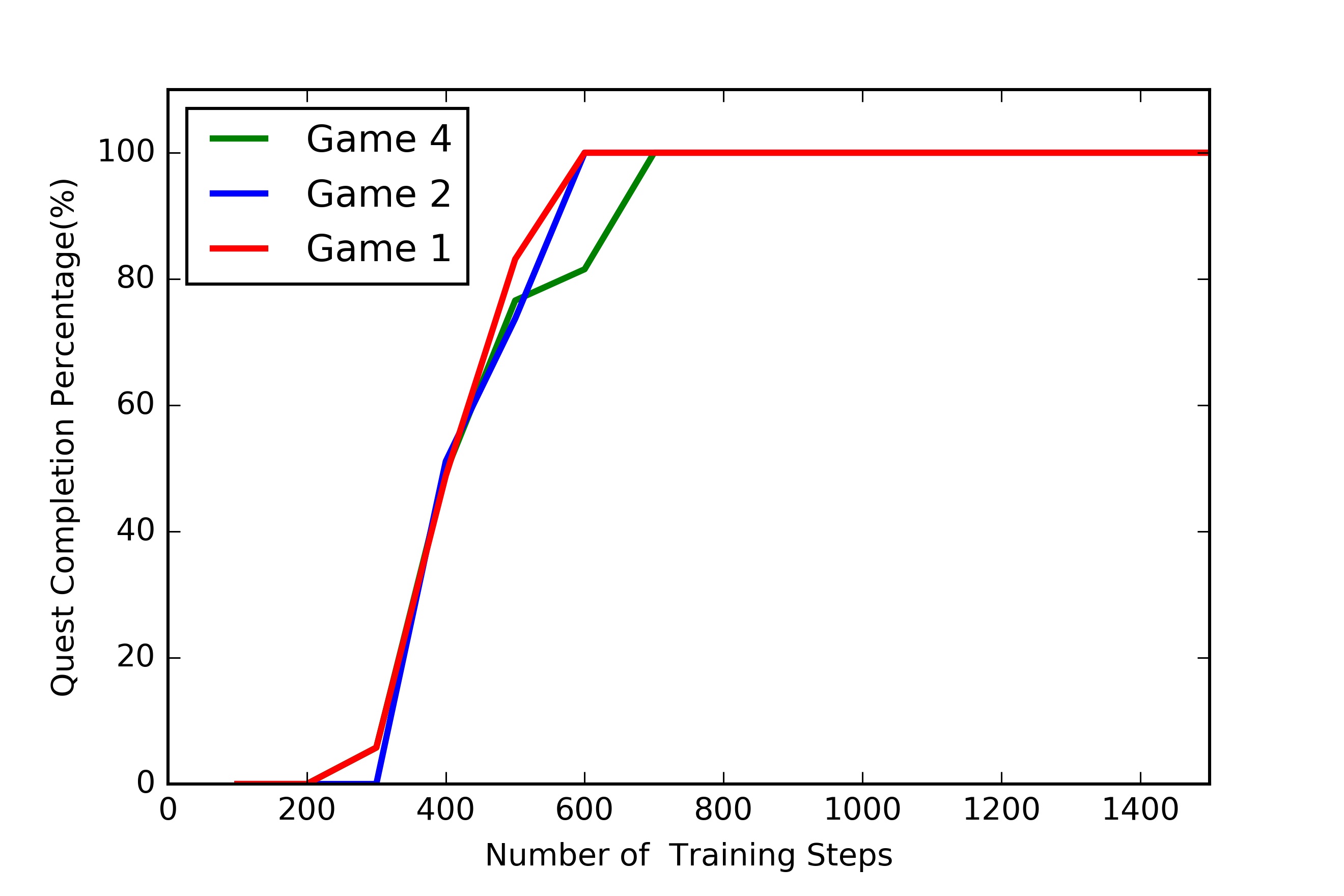}
  \caption{Quest Completion Percentage}
  \label{fig:lsqcomps}
\end{subfigure}
\caption{Training curves of larger multi-task policy distillation agent (student) with $1^{st}$ linear layer size as $100$}
\label{fig:lsperfs}
\end{figure}

From the training curves for Game $4$ in Figure \ref{fig:lsperfs}, we observe that the learning is much more steadier and faster compared to the smaller student network experiments \ref{smallernet}. Making the network wider enhances the modelling capability of the student network. The results in Figure \ref{fig:lsperfs} indicate that, with more modelling power, the student can learn much faster on contrasting domains.

\subsection{What is the mechanics of multi-task policy distillation?}
\label{HeatMaps}
\begin{figure}[htbp]
\begin{subfigure}{0.45\textwidth}
  \centering
  \includegraphics[width = 0.48\textwidth]{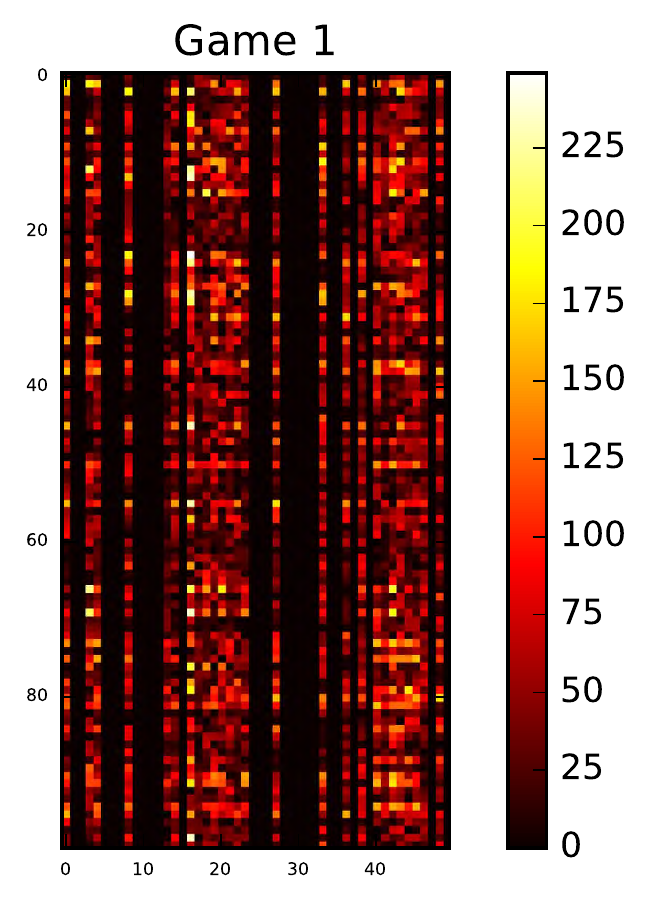}
  \includegraphics[width = 0.48\textwidth]{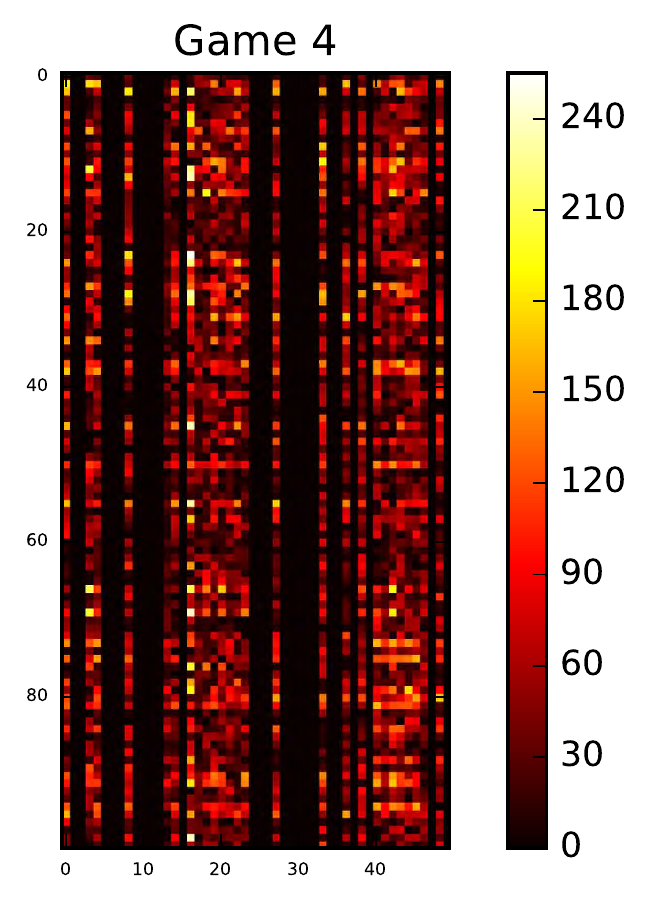}
  \caption{ReLU layer vs mean-pool layer}\label{fig:repr_s}
\end{subfigure}\hfill
\begin{subfigure}{0.2\textwidth}
  \centering
  \includegraphics[width = 0.5\textwidth]{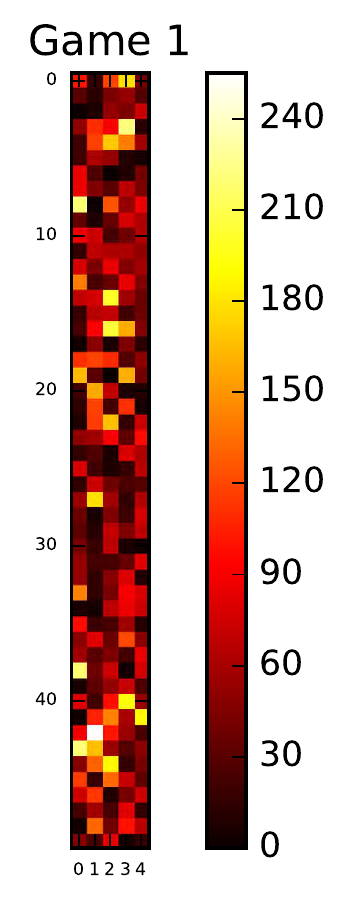}\hspace*{\fill}
  \includegraphics[width = 0.5\textwidth]{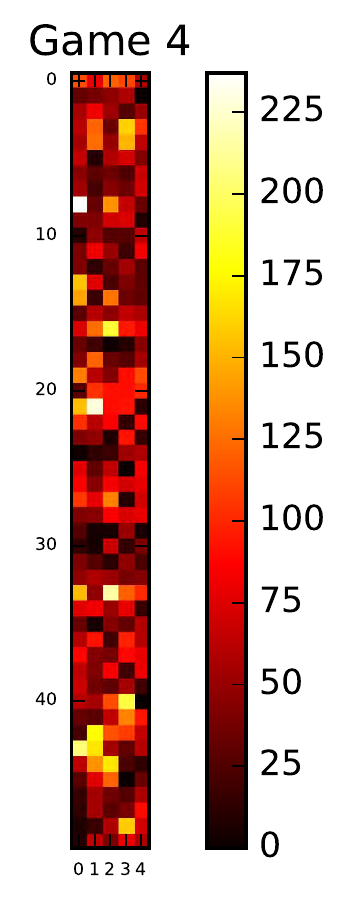}
  \caption{Action value layer vs ReLU layer}\label{fig:act_s}
\end{subfigure}\hspace*{\fill}
\begin{subfigure}{0.2\textwidth}
  \centering
  \includegraphics[width = 0.5\textwidth]{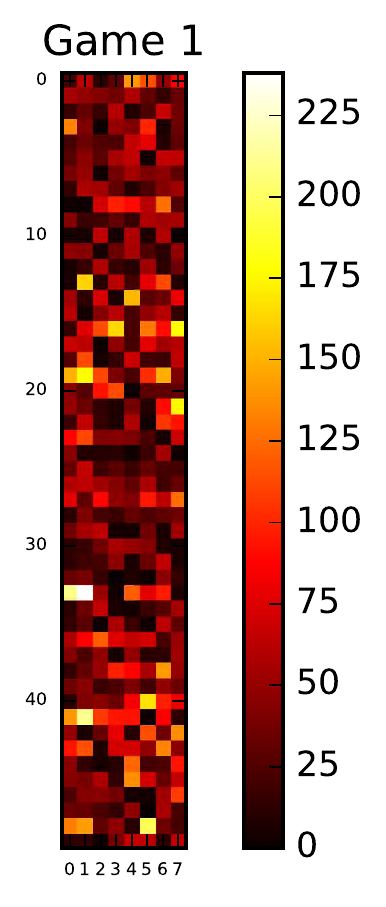}\hspace*{\fill}
  \includegraphics[width = 0.5\textwidth]{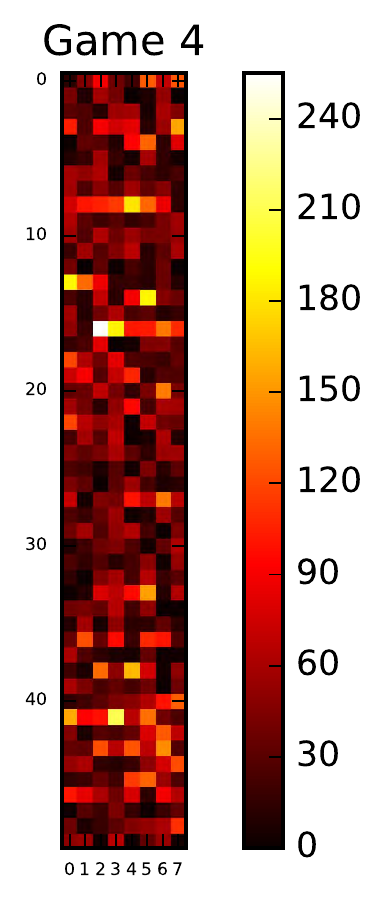}
  \caption{Object value layer vs ReLU layer}\label{fig:obj_s}
\end{subfigure}
\caption{Heat-maps of different layer combinations for \textbf{smaller} student network trained on game $1,2,4$ with size of $1^{st}$ linear layer as \textbf{$50$}. The heat-maps in (a) are very similar for the two games. All the diversity can be seen in the heat-maps in (b) \& (c)}
\label{figure:hmapss}
\end{figure}

\begin{figure}[htbp]
\begin{subfigure}{0.55\textwidth}
  \centering
  \includegraphics[width = 0.48\textwidth]{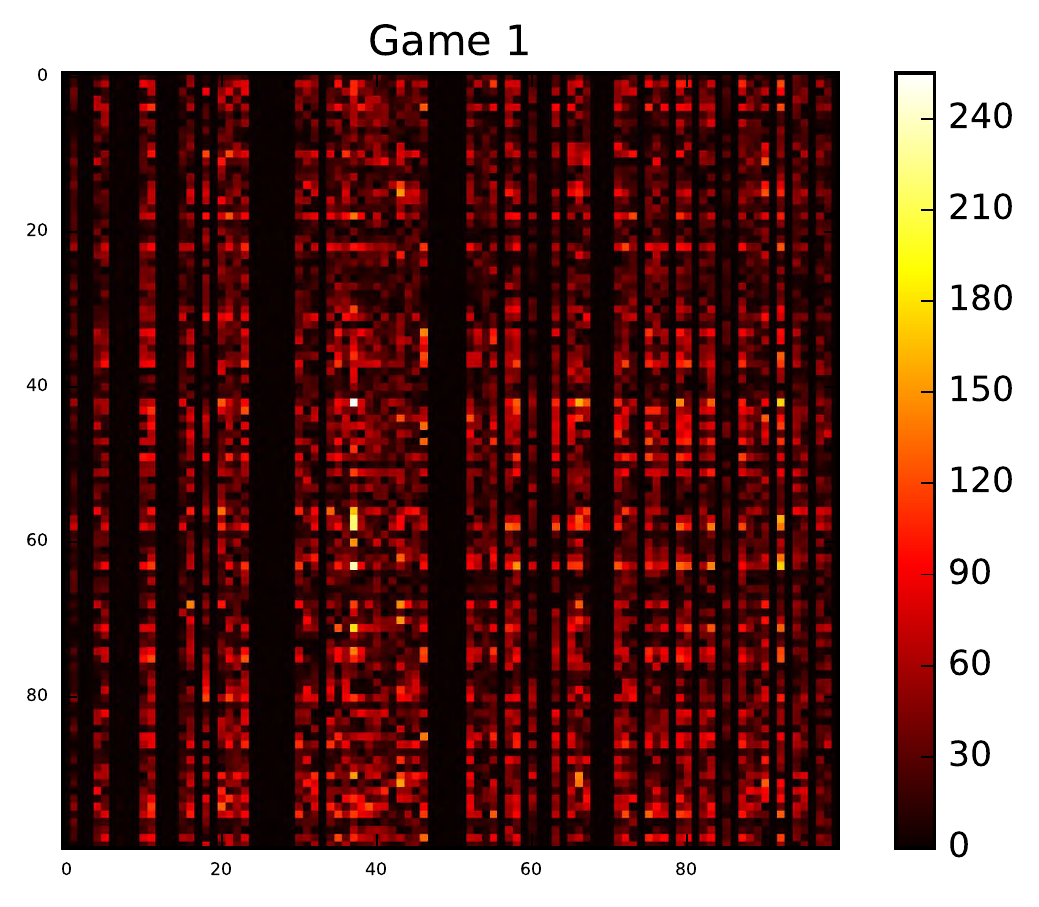}
  \includegraphics[width = 0.48\textwidth]{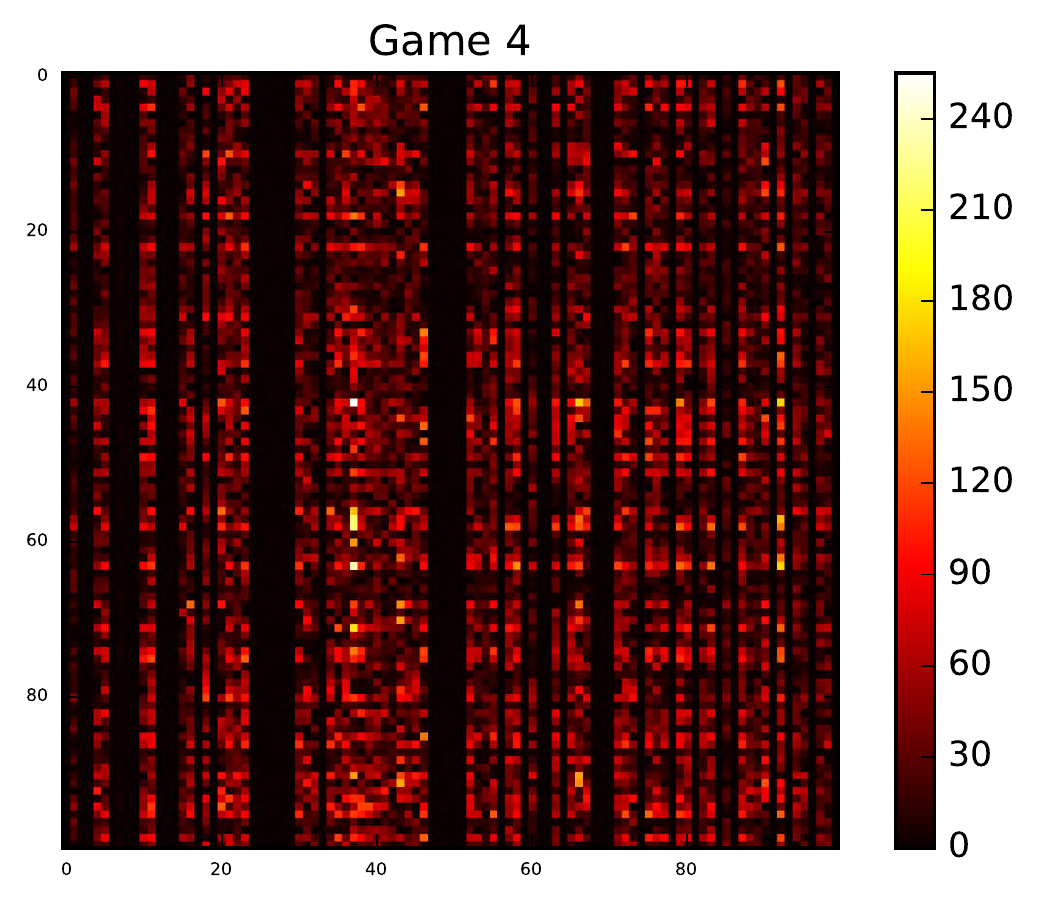}
  \caption{ReLU layer vs mean-pool layer}\label{fig:repr_l}
\end{subfigure}\hfill
\begin{subfigure}{0.2\textwidth}
  \centering
  \includegraphics[width = 0.5\textwidth]{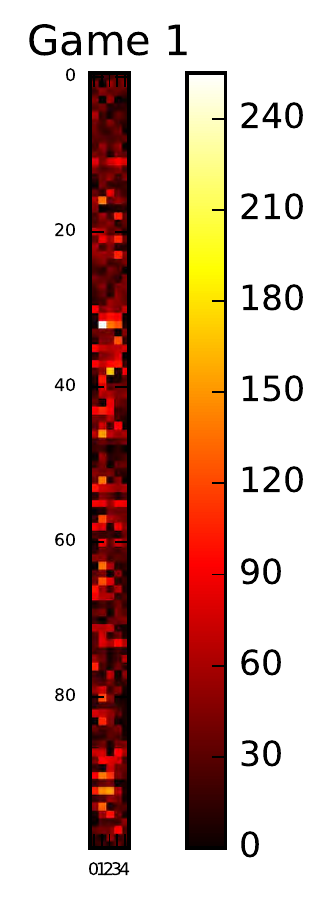}\hspace*{\fill}
  \includegraphics[width = 0.5\textwidth]{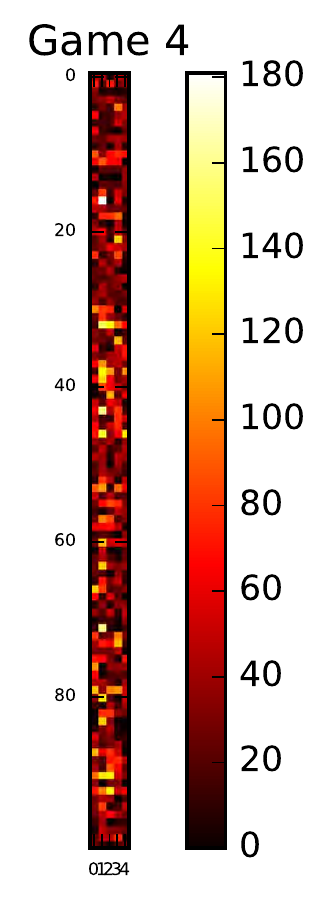}
  \caption{Action value layer vs ReLU layer}\label{fig:act_l}
\end{subfigure}\hspace*{\fill}
\begin{subfigure}{0.2\textwidth}
  \centering
  \includegraphics[width = 0.5\textwidth]{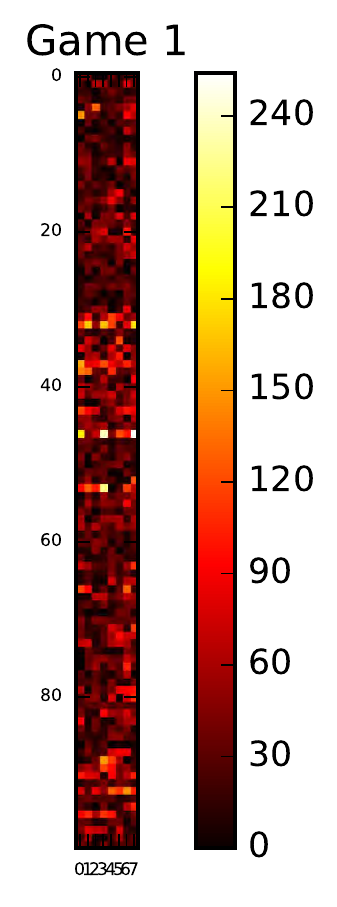}\hspace*{\fill}
  \includegraphics[width = 0.5\textwidth]{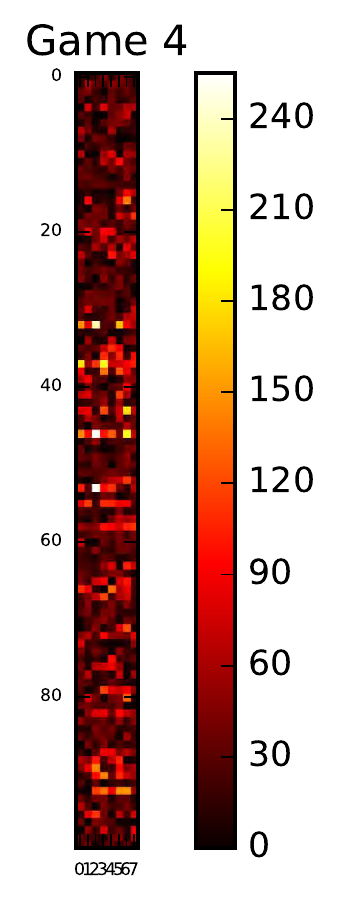}
  \caption{Object value layer vs ReLU layer}\label{fig:obj_l}
\end{subfigure}
\caption{Heat-maps of different layer combinations for \textbf{larger} student network trained on game $1,2,4$ with size of $1^{st}$ linear layer as \textbf{$100$}. The heat-maps in (a) are very similar for the two games. All the diversity can be seen in the heat-maps in (b) \& (c)}
\label{figure:hmapsl}
\end{figure}

Intrigued by the positive results obtained on simply widening a single layer in the student network, we went a step further to interpret the intricacies of the student network. In this experiment, we analyze how inputs from different games affect different parts of the student network. In order to do this, we compare jacobians evaluated for different combinations of network layers for different games. $100$ states are sampled from a single game at once, and the jacobians are calculated with these states as input's to the network. Subsequently, we take an average over the jacobians evaluated over the $100$ sampled states. This process is repeated for the next game and once we have the average jacobian matrices for the required games, we normalize the corresponding jacobian matrices and then scale the absolute values of elements in the resultant matrices to $255$. Final heat-maps are generated for the scaled matrices and can be seen in the Figure \ref{figure:hmapss} \& \ref{figure:hmapsl}.\\
We have chosen the following combinations of layers for evaluating the jacobians for a game $i$, where $i\in \{1,4\}$:
\begin{itemize}
    \item combination 1: jacobian of ReLU layer w.r.t mean-pool layer outputs
    \item combination 2: jacobian of $i^{th}$ action value layer w.r.t ReLU layer
    \item combination 3: jacobian of $i^{th}$ object value layer w.r.t ReLU layer
\end{itemize}
In both Figure \ref{figure:hmapss} \& \ref{figure:hmapsl} we compare the heat-maps between game $1$ \& $4$. We see that in case of both the smaller and larger student network, the heat-maps for combination 1 are very similar where as most of the contrast can be visualized only in the heat-maps for combination 2 \& 3. In the Table \ref{tab:quantifyheat} we numerically quantify the similarities and dissimilarities between the heat-maps. For this we take an average of the absolute values of difference between the maps for game 1 \& 4 for different combinations of network layers.

These results indicate that the learning in the multi-task policy distillation is two-fold. Firstly, a part of the network, the portion between input and ReLU Layer, is specializing to learning something universal and generalizable from different game sources. The results in Section \ref{HeatMaps} motivate us to believe that this portion is responsible for learning generalizable representations from multiple game sources. Secondly, the portions of network between different controllers and the ReLU layer are learning the corresponding game specific information. This especially is the reason why making this part of the network deeper or wider will help get better performance on the complex games as can be seen in Section \ref{larger}.
\begin{table}[H]
  \caption{Average absolute difference between heat-maps of game $1,4$ for different combinations of network layers using the two different sized versions of student network}
  \label{heatmaps}
  \centering
  \begin{tabular}{lll}
    \toprule
    Combination     & $1^{st}$ linear layer size $50$     & $1^{st}$ linear layer size $100$ \\
    \midrule
    ReLU vs mean-pool & 23.58 & 23.11\\
    Action value vs ReLU & 138.02  & 123.56 \\
    Object value vs ReLU     & 124.43 & 129.53\\
    \bottomrule
    \label{tab:quantifyheat}
  \end{tabular}
\end{table}
\subsection{What can we say from the visualization of word embeddings?}
\label{embed}
\begin{figure}[htbp]
\begin{subfigure}{0.32\textwidth}
  \centering
  \includegraphics[width=1\textwidth]{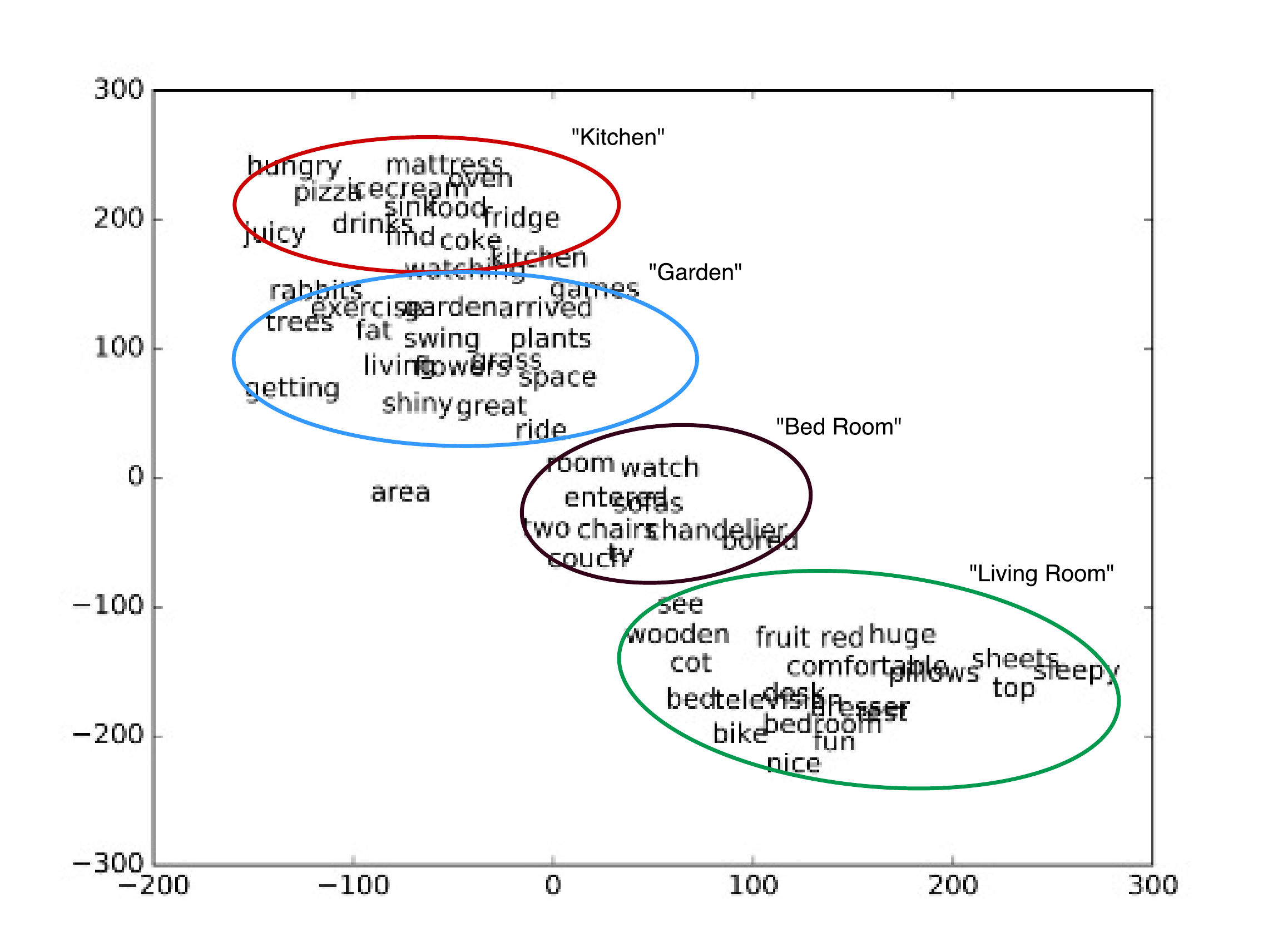}
  \caption{game $1$}
\end{subfigure}\hspace*{\fill}
\begin{subfigure}{0.32\textwidth}
  \centering
  \includegraphics[width=1\textwidth]{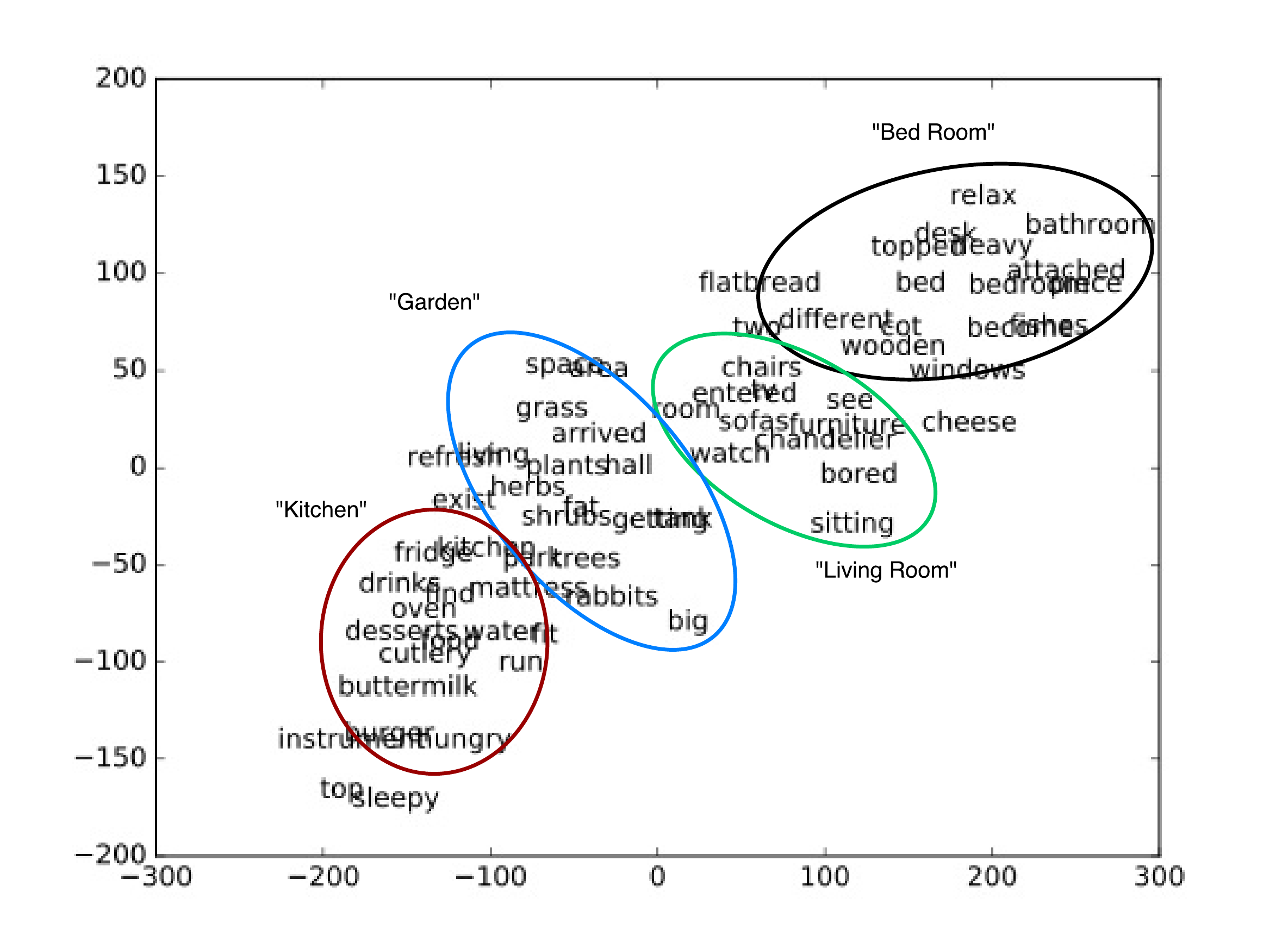}
  \caption{game $2$}
\end{subfigure}\hspace*{\fill}
\begin{subfigure}{0.32\textwidth}
  \centering
  \includegraphics[width=1\textwidth]{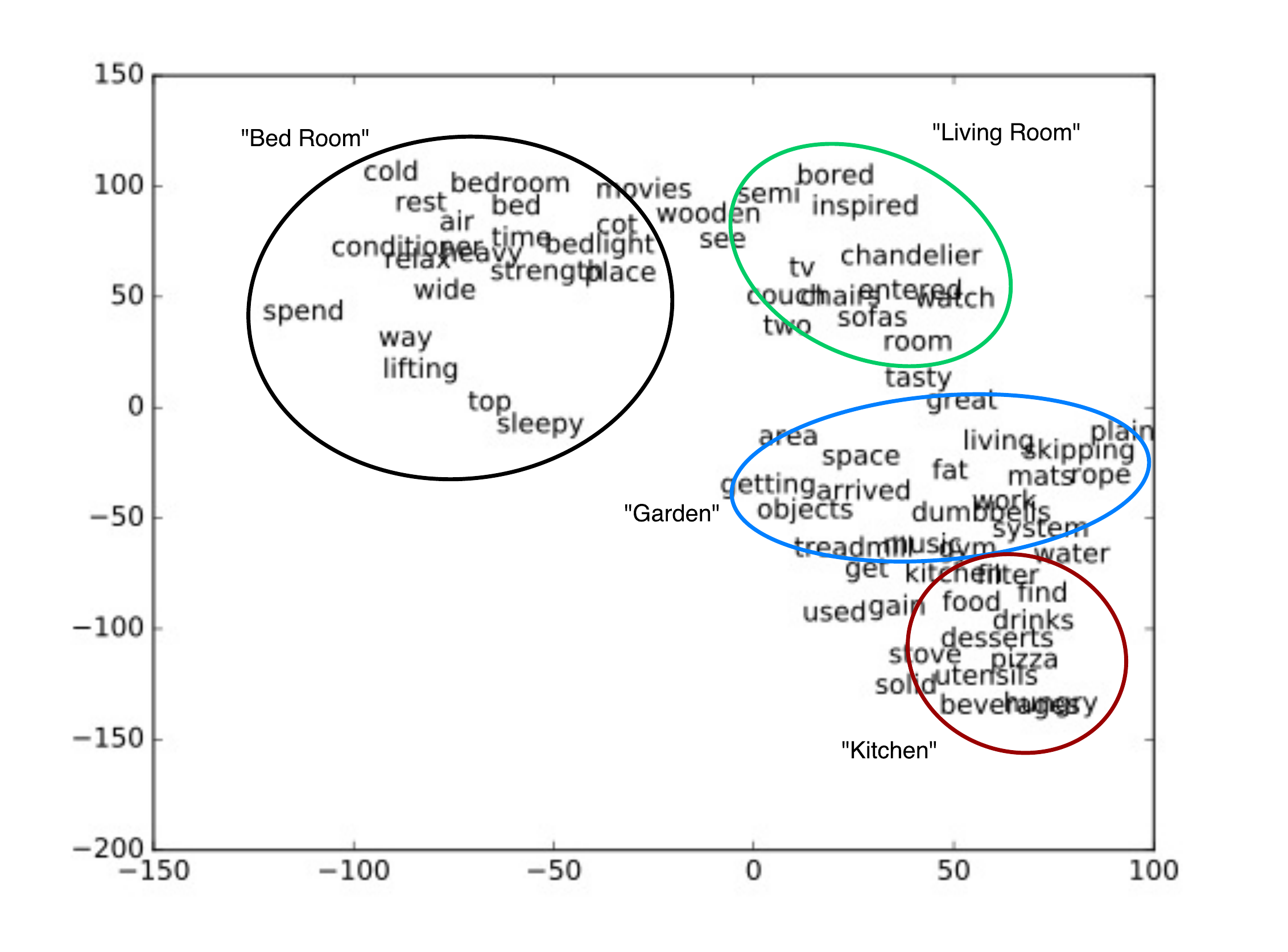}
  \caption{game $3$}
\end{subfigure}
\caption{t-SNE plots of embeddings learnt by multi-task distillation agent (student) trained on games $1,2,3$.}
\label{fig:embedS}
\end{figure}

\begin{figure}[htbp]
\begin{subfigure}{0.32\textwidth}
  \centering
  \includegraphics[width=1\textwidth]{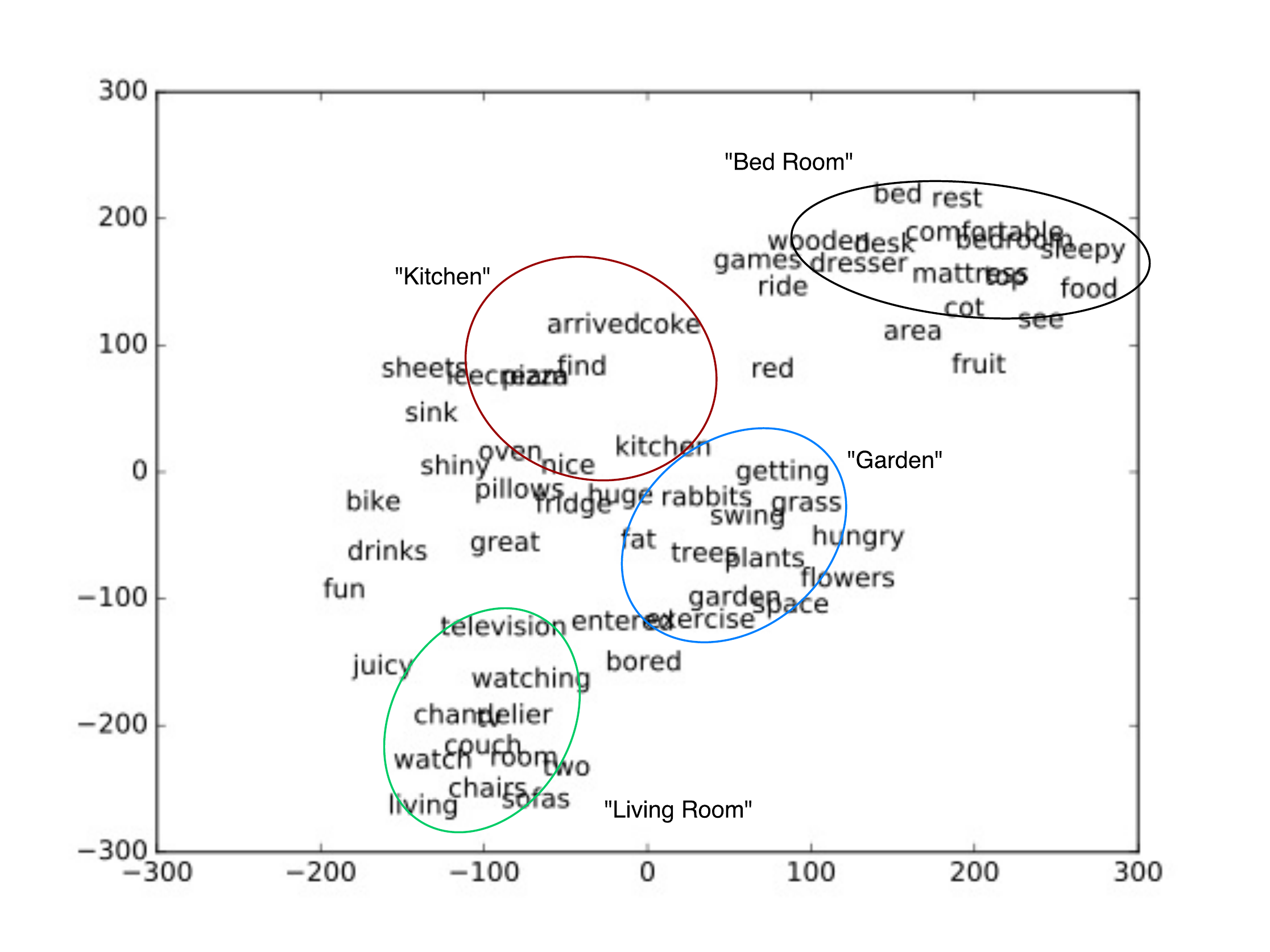}
  \caption{game $1$}
\end{subfigure}\hspace*{\fill}
\begin{subfigure}{0.32\textwidth}
  \centering
  \includegraphics[width=1\textwidth]{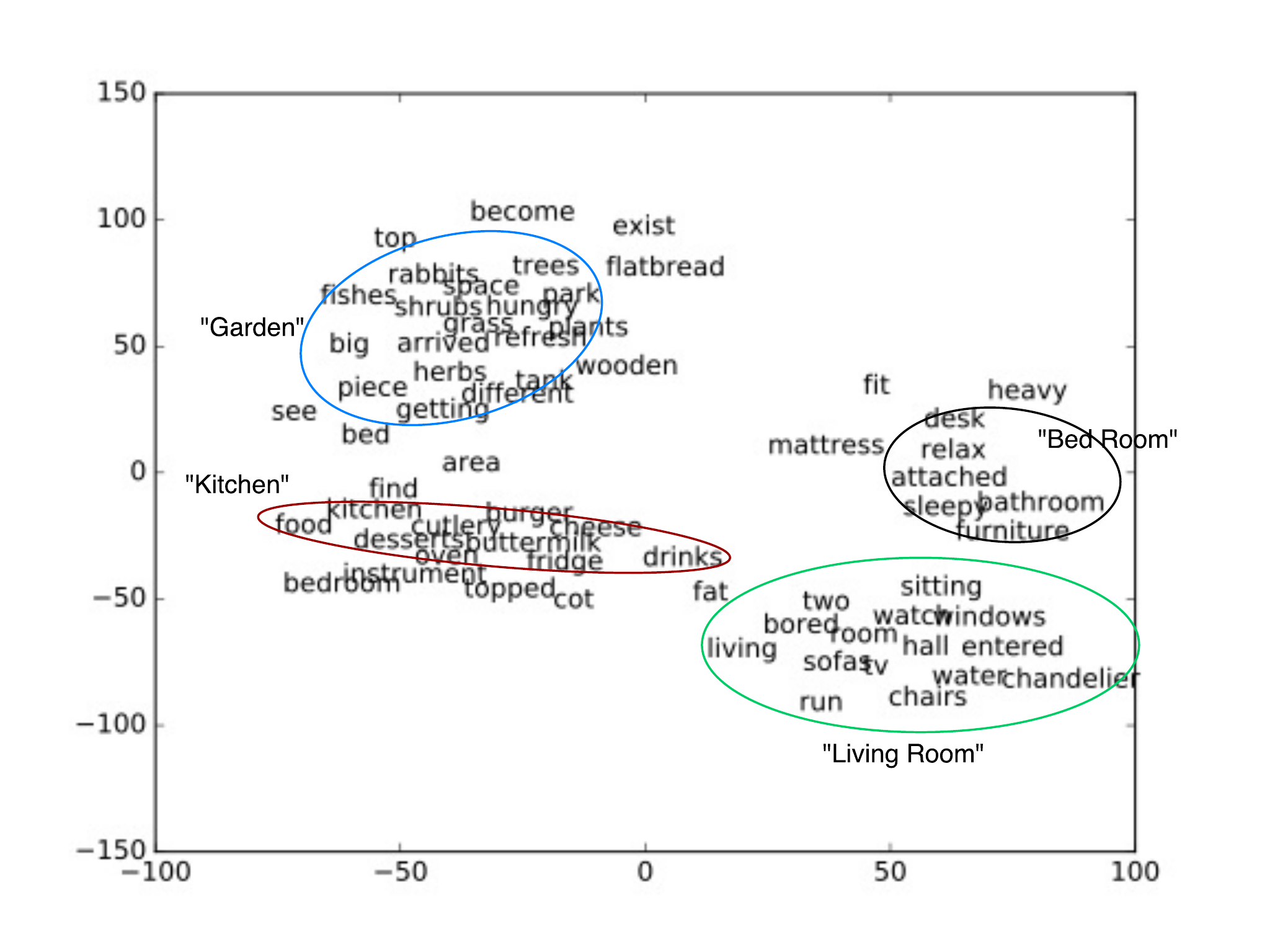}
  \caption{game $2$}
\end{subfigure}\hspace*{\fill}
\begin{subfigure}{0.32\textwidth}
  \centering
  \includegraphics[width=1\textwidth]{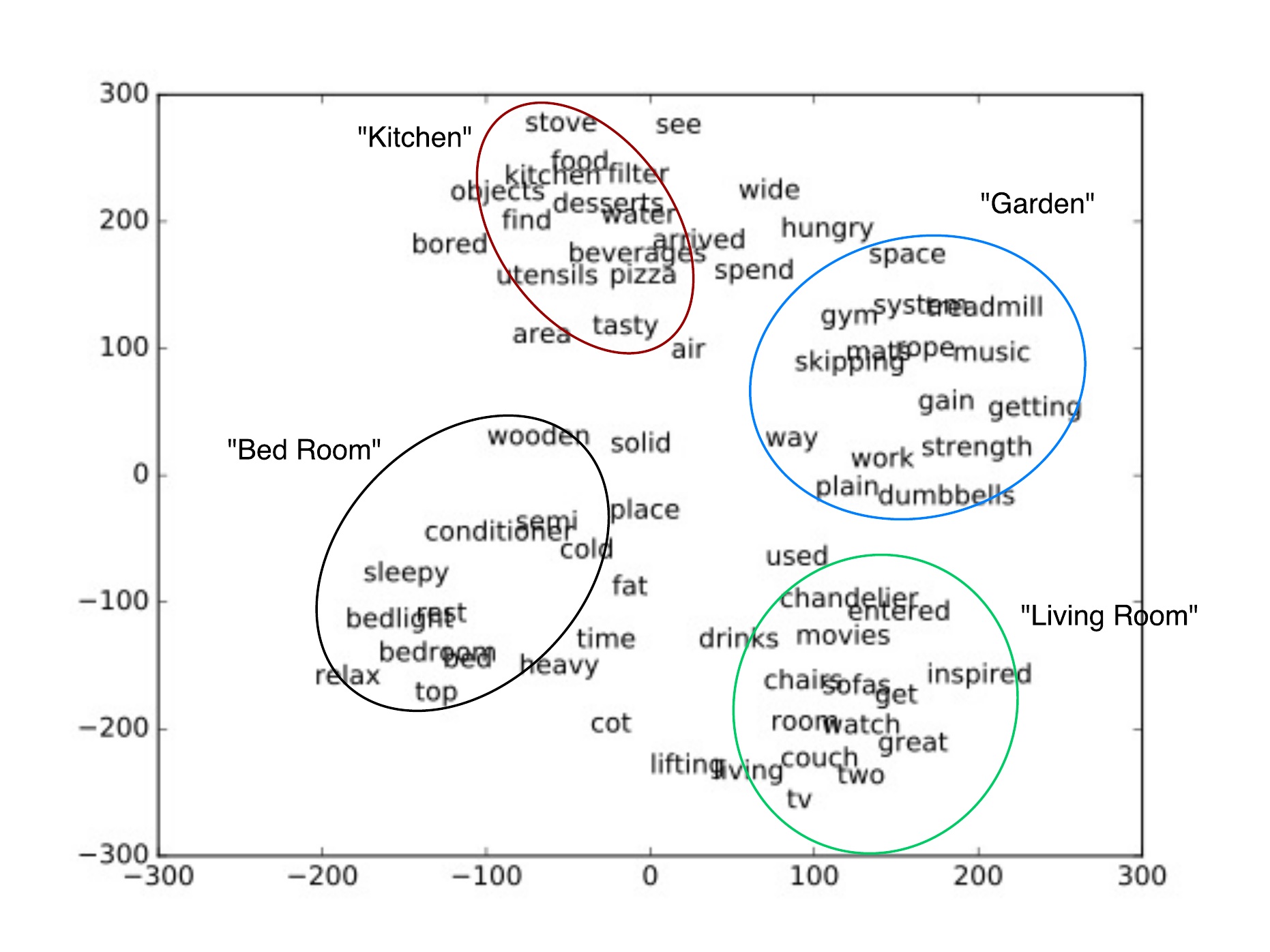}
  \caption{game $3$}
\end{subfigure}
\caption{t-SNE plots of embeddings learnt by LSTM-DQN teachers trained on games $1,2,3$ respectively.}
\label{fig:embedT}
\end{figure}

In this section we analyze the t-SNE \cite{tsne} embeddings learnt by the multi-task policy distillation method on learning games 1,2 and 3 combined and compare them with the embeddings learnt by corresponding single game teachers. We train a new student network for this experiment. Games 1,2, and 3 all have the same layout but have different vocabularies. For this we first passed every word in the vocabulary of a game through the LSTM layer and took its output and then projected it onto a 2 dimensional space using t-SNE algorithm \cite{tsne} and plotted the representations. This was done for all the teacher networks and the student network. In this experiment we used a student network of same size as that of corresponding teacher networks. From the Figure \ref{fig:embedS} and Figure \ref{fig:embedT}, we can infer that the word embeddings learnt by a single student network are much more compact and better clustered as compared to those learnt by multiple single game teachers. Also, the number of outliers in the clusters are much less in the case of student embeddings as can be seen in the Figure \ref{fig:embedS}.

We believe that the student network is able to associate words with the rooms and the tasks (quests) better than the teacher networks, by benefiting from textual descriptions from different games.\\
One more factor that could enable student to learn better language representations is that, in the process of policy distillation only the final optimal policies of the teacher are learnt by the student network. These stable targets to the student network could help the network to learn better, unlike the case of LSTM-DQN teachers training where the network has to represent a series of policies that are obtained in the policy iteration cycle of the Q-learning Algorithm.

\subsection{How to do transfer learning using policy distillation?}
\label{transfer}
Going a step further in comparing the utility of representations learnt by the student with that of a single game teacher, we created game $5$ which has parts of its vocabulary common with each of the games $1,2,3$. We used single game teachers $T_1$,$T_2$,$T_3$ trained on games $1,2,3$ respectively. The student which we will denote by $S$ is trained using these expert teachers $T_1$,$T_2$,$T_3$. For this experiment we take the Game 5 and train $4$ agents $A_1,A_2,A_3,A_4$ by fixing the embeddings of possible words with the embeddings learnt by $T_1$,$T_2$,$T_3$ and $S$ respectively. Here, by possible words we mean those words that are common between the already learnt network's game source and game-$5$. The transferred embeddings are fixed and are not updated during training. The embeddings for the words that are not common are initialized randomly just as in the case of a normal LSTM-DQN teacher.

We also train two more agents, one is $A_5$ that has all the word embeddings initialized randomly, and other is $A_6$ which has all the possible word embeddings from game-5 initialized using all possible embeddings from teachers $T_1$,$T_2$,$T_3$ at the same time. In the case of $A_6$ when there is more than one possible embedding source for a word, the source is chosen randomly. Agent $A_5$ an $A_6$ serve as baselines for this experiment.

\begin{figure}[htbp]
\centering
\begin{subfigure}{.5\textwidth}
  \centering
  \includegraphics[width=0.9\linewidth]{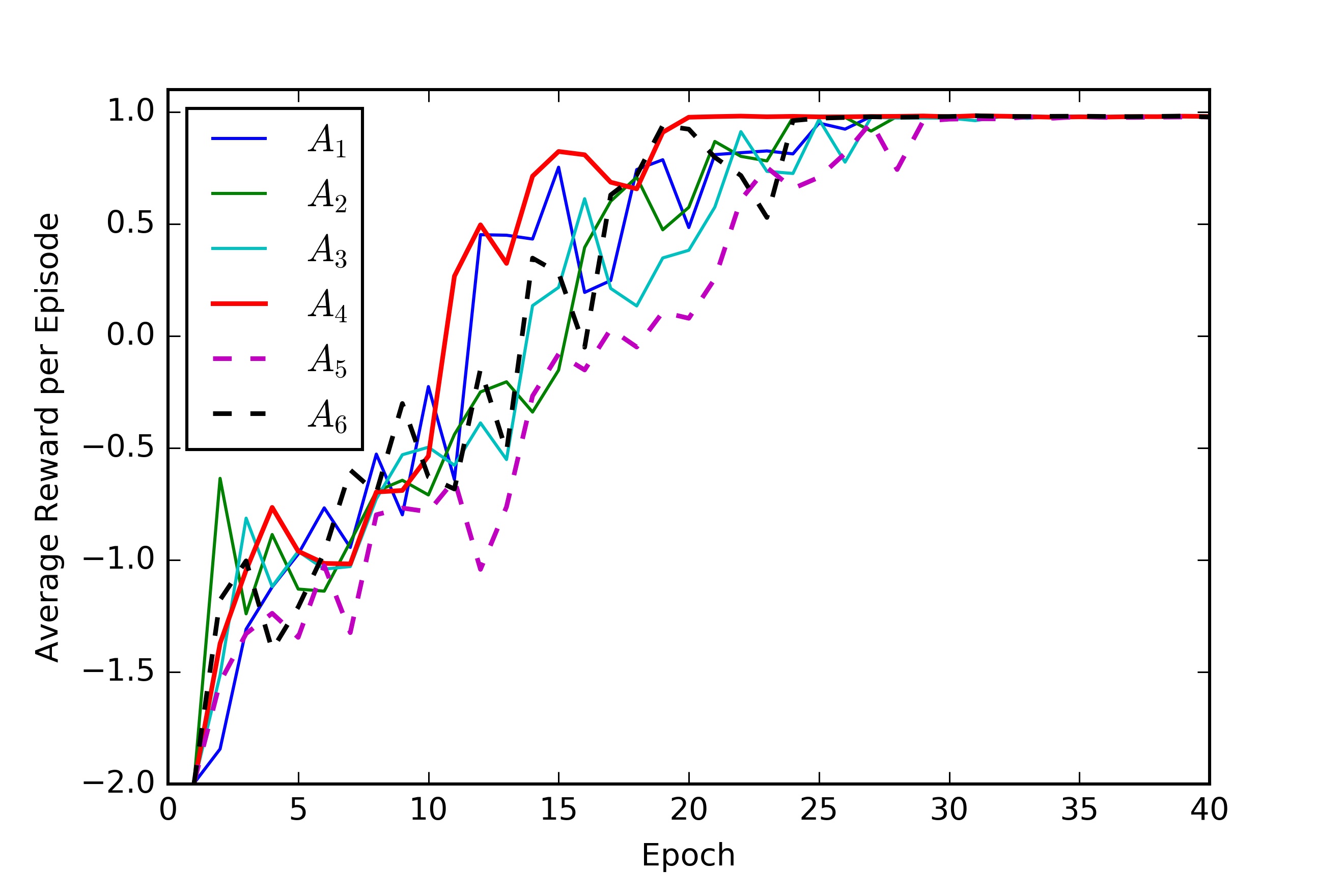}
  \caption{Average reward per episode}
\end{subfigure}%
\begin{subfigure}{.5\textwidth}
  \centering
  \includegraphics[width=0.9\linewidth]{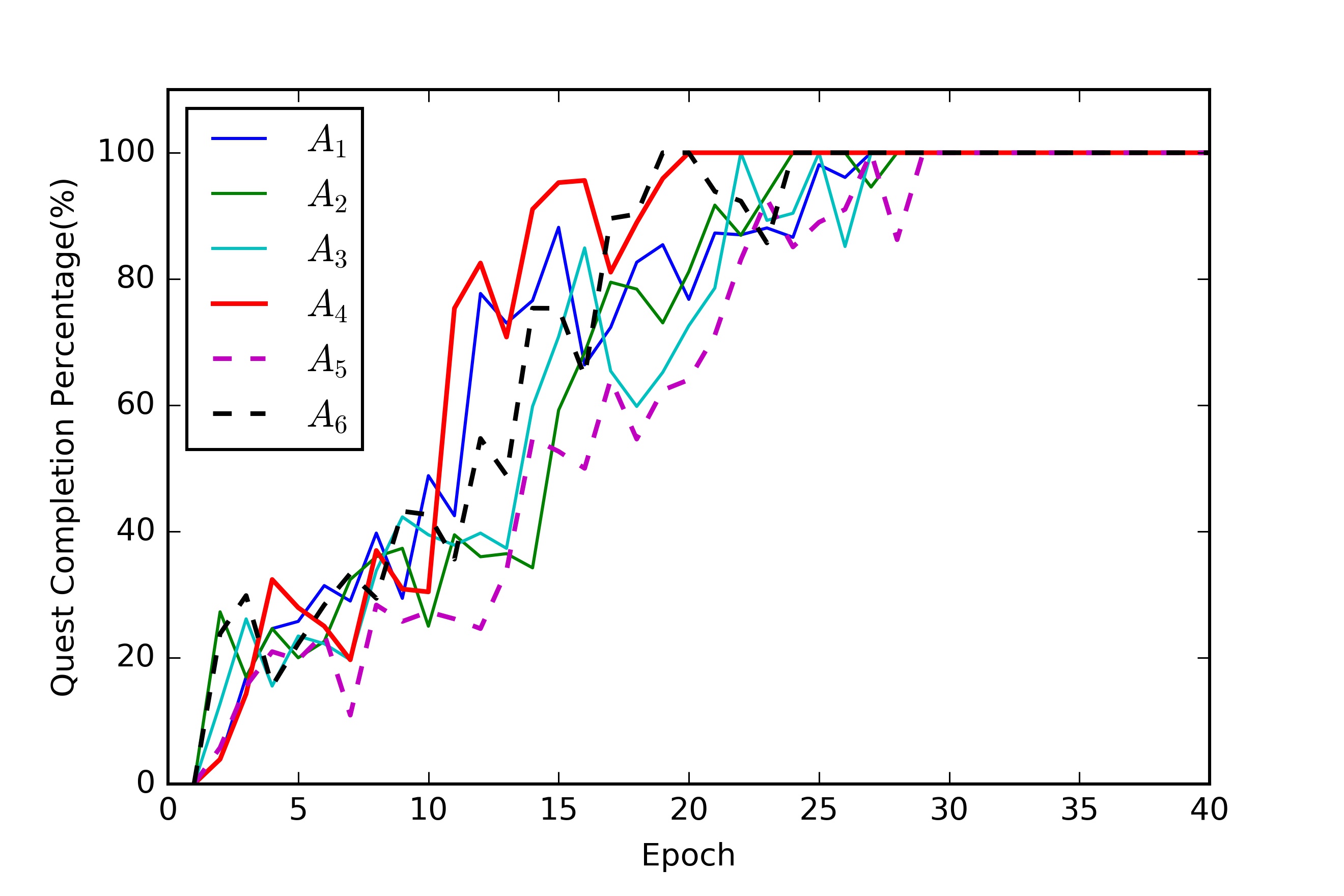}
  \caption{Quest completion percentage}
\end{subfigure}
\caption{Training curves for LSTM-DQN agents $A_1,A_2,A_3,A_4,A_5,A_6$ training on game $5$. Agent $A_1,A_2,A_3,$ are initialized with learnt embeddings from LSTM-DQN teachers $T_1$,$T_2$,$T_3$ respectively. Agent $A_4$ is initialized with learnt embeddings from student $S$. Agent $A_5$ has randomly initialized embeddings. Agent $A_6$ has all possible embeddings initialized from all the LSTM-DQN teachers $T_1$,$T_2$,$T_3$ combined}
\label{fig:transfer}
\end{figure}

From the Figure \ref{fig:transfer} we infer that the teacher $A_{4}$ that had embeddings initialized from the student $S$ is able to learn much faster than the others. Most importantly, we see that it performs better than $A_6$, which is the usual method for language expansion. This strengthens our belief that student is able to learn useful representations of language from multiple game sources.

\subsection{How is the performance of policy distillation when compared to multi-task LSTM-DQN?}
\label{mlstmdqn}
\begin{figure}[htp]
\centering
\begin{subfigure}{.5\textwidth}
  \centering
  \includegraphics[width=0.9\linewidth]{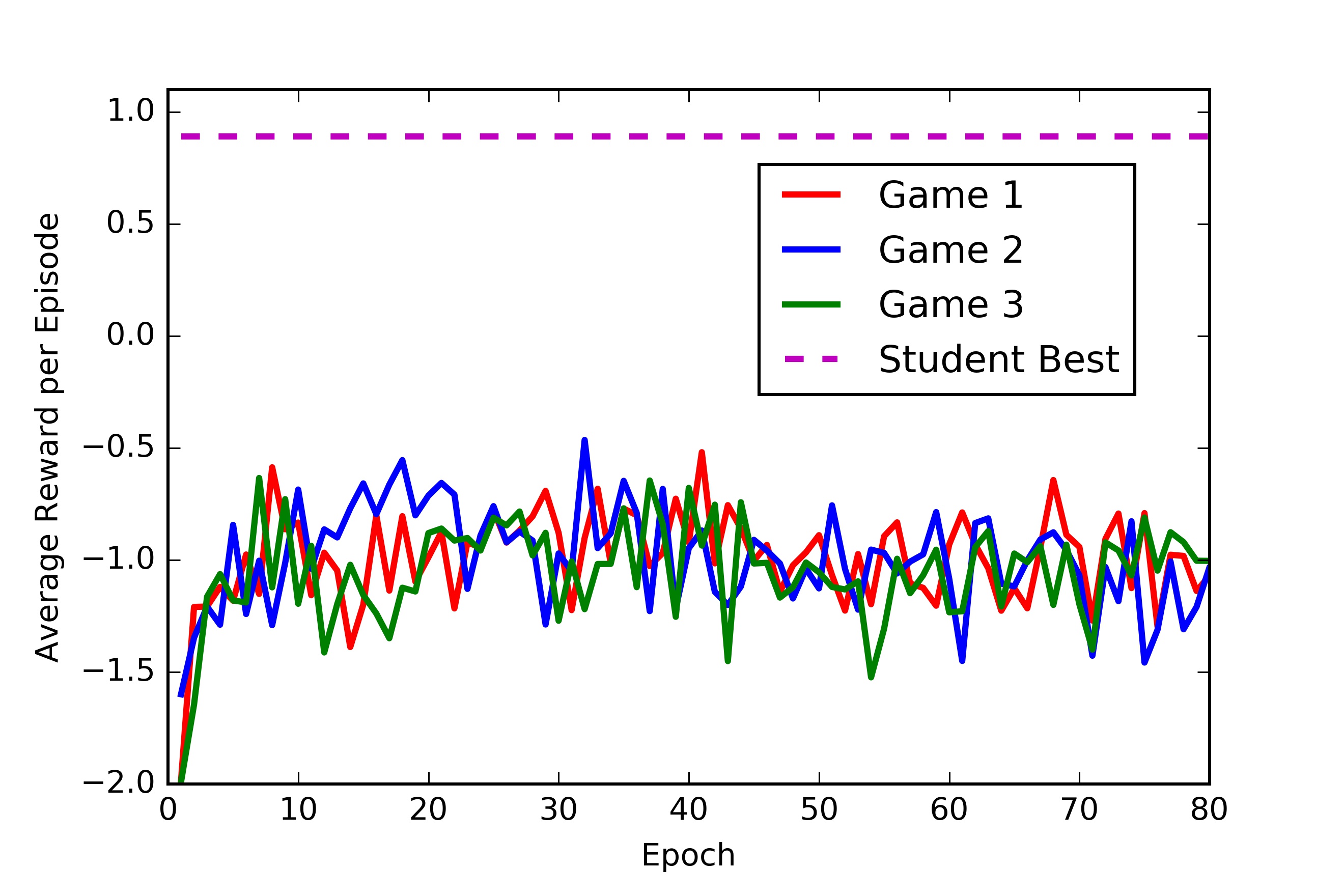}
  \caption{Average reward per episode}
\end{subfigure}%
\begin{subfigure}{.5\textwidth}
  \centering
  \includegraphics[width=0.9\linewidth]{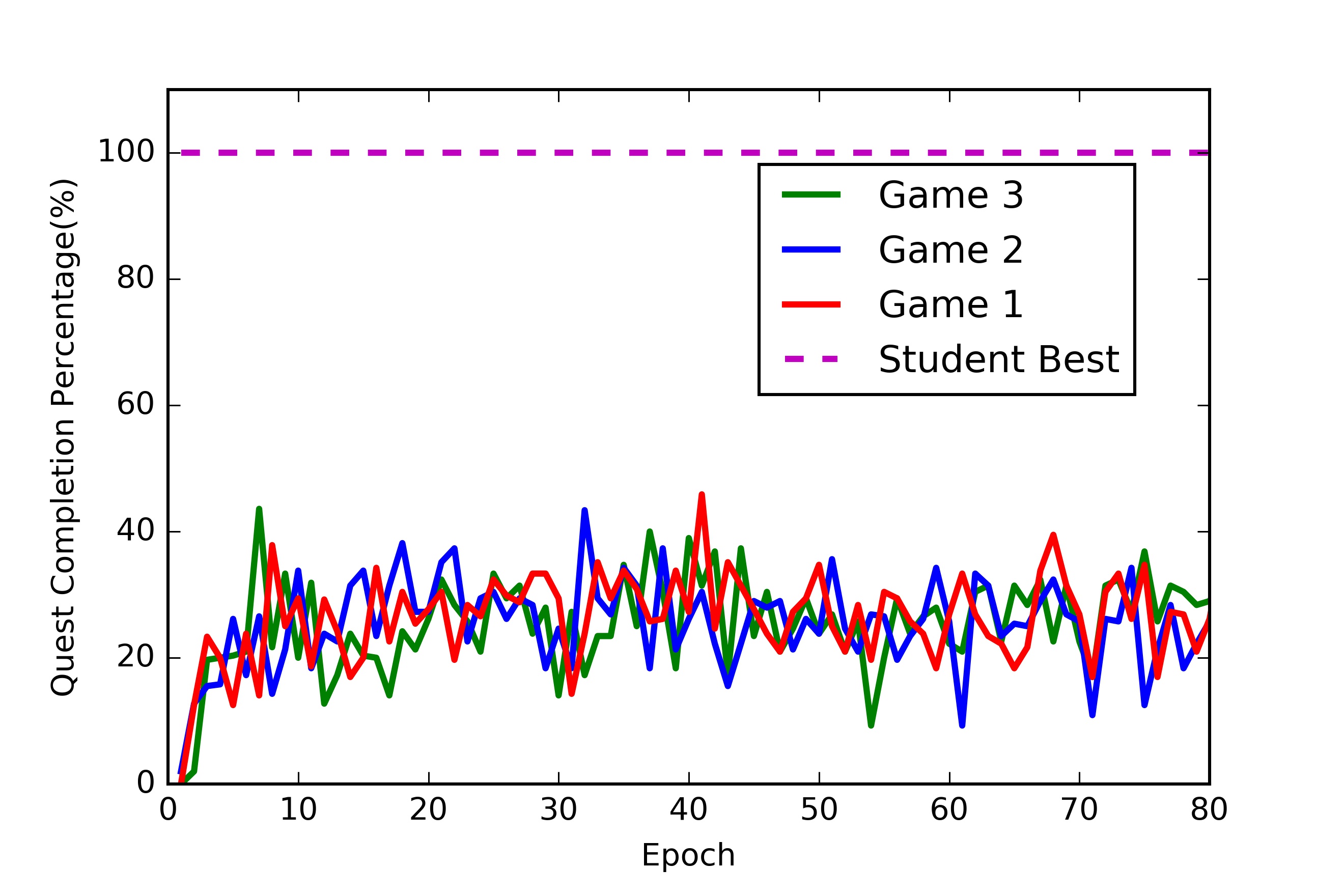}
  \caption{Quest completion percentage}
\end{subfigure}
\caption{Training curves for multi-task LSTM-DQN}
\label{fig:mldqn}
\end{figure}
In this section we analyze the performance of the multi-task LSTM-DQN. We trained a multi-task LSTM-DQN agent on games $1,2,3$. The performance of this method is very bad as compared to the multi-task policy distillation method as can be seen in Figure \ref{fig:mldqn}. The multi task LSTM-DQN agent is able to solve only about $30\%$ of the quests on an average, given that the episodes are ended after $20$ steps, whereas multi-task policy distillation was able to solve $100\%$ of the quests. Even the average reward per episode obtained is much less than the best possible reward of $0.98$.

Analysis of the embeddings learnt by the agent for words in vocabulary of games $1,2,3$ showed that the embeddings learnt by this method are almost random. Even the semantically similar words are not clustering close to each other. Even though we have a set of text descriptions from different game sources, this method fails to generate robust embeddings. This could happen because the errors are not propagating steadily through the representation layer of the network, as the policy that the agent is learning keeps changing due to the policy iteration cycle of the Q-Learning algorithm \cite{qlearning}. Since, there are 3 sets of games we are learning here, the effective targets to the representation layer part of the network are even more unstable. This result agrees with similar result for convolutional DQN model in \cite{DBLP:distillation}.

\section{Conclusion}

In this paper, we applied the recently proposed policy distillation method \cite{DBLP:distillation} to text-based games. Our experiments verify that policy distillation is an effective approach even for LSTM-DQN where each game might have different vocabulary. We also provided an extensive analysis on how will distillation work when games have contradicting state dynamics. We provide visualizations that shed some light on the internal working of policy distillation agents. It will be interesting to see if the same analysis will hold for games with larger vocabulary space. Finally, we believe that this is an important step in the direction of designing agents that can expand their language in a completely online fashion (aka continual learning).

\nocite{he2016deep}
\nocite{nogueira2016end}
\nocite{sukhbaatar2015mazebase}
\bibliographystyle{plainnat}
\bibliography{references}

\begin{thebibliography}{17}
\providecommand{\natexlab}[1]{#1}
\providecommand{\url}[1]{\texttt{#1}}
\expandafter\ifx\csname urlstyle\endcsname\relax
  \providecommand{\doi}[1]{doi: #1}\else
  \providecommand{\doi}{doi: \begingroup \urlstyle{rm}\Url}\fi

\bibitem[Bengio et~al.(2007)Bengio, Lamblin, Popovici, Larochelle,
  et~al.]{bengio2007greedy}
Yoshua Bengio, Pascal Lamblin, Dan Popovici, Hugo Larochelle, et~al.
\newblock Greedy layer-wise training of deep networks.
\newblock 2007.

\bibitem[He et~al.(2015)He, Chen, He, Gao, Li, Deng, and Ostendorf]{he2015deep}
Ji~He, Jianshu Chen, Xiaodong He, Jianfeng Gao, Lihong Li, Li~Deng, and Mari
  Ostendorf.
\newblock Deep reinforcement learning with a natural language action space.
\newblock \emph{arXiv preprint arXiv:1511.04636}, 2015.

\bibitem[He et~al.(2016)He, Ostendorf, He, Chen, Gao, Li, and Deng]{he2016deep}
Ji~He, Mari Ostendorf, Xiaodong He, Jianshu Chen, Jianfeng Gao, Lihong Li, and
  Li~Deng.
\newblock Deep reinforcement learning with a combinatorial action space for
  predicting and tracking popular discussion threads.
\newblock \emph{arXiv preprint arXiv:1606.03667}, 2016.

\bibitem[Hinton et~al.(2006)Hinton, Osindero, and Teh]{hinton2006fast}
Geoffrey~E Hinton, Simon Osindero, and Yee-Whye Teh.
\newblock A fast learning algorithm for deep belief nets.
\newblock \emph{Neural computation}, 18\penalty0 (7):\penalty0 1527--1554,
  2006.

\bibitem[Maaten and Hinton(2008)]{tsne}
Laurens van~der Maaten and Geoffrey Hinton.
\newblock Visualizing data using t-sne.
\newblock \emph{Journal of Machine Learning Research}, 9\penalty0
  (Nov):\penalty0 2579--2605, 2008.

\bibitem[Mikolov et~al.(2013)Mikolov, Sutskever, Chen, Corrado, and
  Dean]{mikolov2013distributed}
Tomas Mikolov, Ilya Sutskever, Kai Chen, Greg~S Corrado, and Jeff Dean.
\newblock Distributed representations of words and phrases and their
  compositionality.
\newblock In \emph{Advances in neural information processing systems}, pages
  3111--3119, 2013.

\bibitem[Mnih et~al.(2015)Mnih, Kavukcuoglu, Silver, Rusu, Veness, Bellemare,
  Graves, Riedmiller, Fidjeland, Ostrovski, et~al.]{dqnnature}
Volodymyr Mnih, Koray Kavukcuoglu, David Silver, Andrei~A Rusu, Joel Veness,
  Marc~G Bellemare, Alex Graves, Martin Riedmiller, Andreas~K Fidjeland, Georg
  Ostrovski, et~al.
\newblock Human-level control through deep reinforcement learning.
\newblock \emph{Nature}, 518\penalty0 (7540):\penalty0 529--533, 2015.

\bibitem[Narasimhan et~al.(2015)Narasimhan, Kulkarni, and Barzilay]{lstmdqn}
Karthik Narasimhan, Tejas~D. Kulkarni, and Regina Barzilay.
\newblock Language understanding for text-based games using deep reinforcement
  learning.
\newblock \emph{CoRR}, abs/1506.08941, 2015.
\newblock URL \url{http://arxiv.org/abs/1506.08941}.

\bibitem[Nogueira and Cho(2016)]{nogueira2016end}
Rodrigo Nogueira and Kyunghyun Cho.
\newblock End-to-end goal-driven web navigation.
\newblock \emph{arXiv preprint arXiv:1602.02261}, 2016.

\bibitem[Pennington et~al.()Pennington, Socher, and
  Manning]{pennington2014glove}
Jeffrey Pennington, Richard Socher, and Christopher~D Manning.
\newblock Glove: Global vectors for word representation.

\bibitem[Rusu et~al.(2015)Rusu, Colmenarejo, G{\"{u}}l{\c{c}}ehre, Desjardins,
  Kirkpatrick, Pascanu, Mnih, Kavukcuoglu, and Hadsell]{DBLP:distillation}
Andrei~A. Rusu, Sergio~Gomez Colmenarejo, {\c{C}}aglar G{\"{u}}l{\c{c}}ehre,
  Guillaume Desjardins, James Kirkpatrick, Razvan Pascanu, Volodymyr Mnih,
  Koray Kavukcuoglu, and Raia Hadsell.
\newblock Policy distillation.
\newblock \emph{CoRR}, abs/1511.06295, 2015.
\newblock URL \url{http://arxiv.org/abs/1511.06295}.

\bibitem[Schaul et~al.(2015)Schaul, Quan, Antonoglou, and
  Silver]{schaul2015prioritized}
Tom Schaul, John Quan, Ioannis Antonoglou, and David Silver.
\newblock Prioritized experience replay.
\newblock \emph{arXiv preprint arXiv:1511.05952}, 2015.

\bibitem[Sukhbaatar et~al.(2015)Sukhbaatar, Szlam, Synnaeve, Chintala, and
  Fergus]{sukhbaatar2015mazebase}
Sainbayar Sukhbaatar, Arthur Szlam, Gabriel Synnaeve, Soumith Chintala, and Rob
  Fergus.
\newblock Mazebase: A sandbox for learning from games.
\newblock \emph{arXiv preprint arXiv:1511.07401}, 2015.

\bibitem[Van~Hasselt et~al.()Van~Hasselt, Guez, and Silver]{van2015deep}
Hado Van~Hasselt, Arthur Guez, and David Silver.
\newblock Deep reinforcement learning with double q-learning.

\bibitem[Wang et~al.(2015)Wang, de~Freitas, and Lanctot]{wang2015dueling}
Ziyu Wang, Nando de~Freitas, and Marc Lanctot.
\newblock Dueling network architectures for deep reinforcement learning.
\newblock \emph{arXiv preprint arXiv:1511.06581}, 2015.

\bibitem[Watkins and Dayan(1992)]{qlearning}
Christopher~JCH Watkins and Peter Dayan.
\newblock Q-learning.
\newblock \emph{Machine learning}, 8\penalty0 (3-4):\penalty0 279--292, 1992.

\bibitem[Weston(2016)]{weston2016dialog}
Jason Weston.
\newblock Dialog-based language learning.
\newblock \emph{arXiv preprint arXiv:1604.06045}, 2016.

\end{thebibliography}
\small

\end{document}